\pdfoutput=1

\documentclass[11pt]{article}

\usepackage{EMNLP2022}

\usepackage{times}
\usepackage{latexsym}

\usepackage[T1]{fontenc}

\usepackage[utf8]{inputenc}

\usepackage{microtype}

\usepackage{inconsolata}

\usepackage{graphicx}
\usepackage{array,multirow}
\usepackage{amsfonts}
\usepackage{amsmath}
\usepackage{xcolor}

\usepackage{bm}
\usepackage{tabularx}
\newcolumntype{L}{>{\raggedright\arraybackslash}X}
\usepackage{fixltx2e}
\usepackage{soul}
\usepackage{booktabs}
\usepackage{hyperref}
\usepackage{soul}
\usepackage{colortbl}
\newcolumntype{C}[1]{>{\centering\arraybackslash}p{#1}}
\newcommand{\stdv}[1]{\scriptsize$\pm$#1}

\title{Semantic Novelty Detection and Characterization in Factual Text \\ Involving Named Entities}

\author{Nianzu Ma$^{\dagger}$,~ Sahisnu Mazumder$^{\ddagger}$,~ Alexander Politowicz$^{\dagger}$, \\ ~ {\bf Bing Liu}$^{\dagger}${\bf,~} {\bf Eric Robertson}$^{\natural}${\bf,~} {\bf Scott Grigsby}$^{\natural}$ \\
  $^{\dagger}$Department of Computer Science, University of Illinois at Chicago, USA \\
  $^{\ddagger}$Intel Labs, USA \\
  $^{\natural}$PAR Government Systems Corporation, USA \\
  \texttt{jingyima005@gmail.com,~sahisnumazumder@gmail.com,~politow2@uic.edu} \\
  \texttt{liub@uic.edu,~\{eric\_robertson,Scott\_Grigsby\}@partech.com} \\
}

\begin{document}
\maketitle
\begin{abstract}
Much of the existing work on text novelty detection has been studied at the topic level, i.e., identifying whether the topic of a document or a sentence is novel or not. Little work has been done at the fine-grained \textit{semantic level} (or \textit{contextual level}). For example, given that we know Elon Musk is the CEO of a technology company, the sentence ``\textit{Elon Musk acted in the sitcom \textit{The Big Bang Theory}}'' is novel and surprising because normally a CEO would not be an actor. Existing topic-based novelty detection methods work poorly on this problem because they do not perform semantic reasoning involving relations between named entities in the text and their background knowledge. This paper proposes an effective model (called PAT-SND) to solve the problem, which can also characterize the novelty. An annotated dataset is also created. Evaluation shows that PAT-SND outperforms 10 baselines by large margins.
\end{abstract}

\section{Introduction}
\label{sec:intro}

Novelty/anomaly detection has been an active research area for decades~\cite{grubbs1969procedures,Chalapathy2018anomaly,pang2020deep}. Recently, it has received increased attention in NLP. Broadly speaking, there are two main types: \textbf{(1)} \textbf{Topic-based novelty detection}, which classifies a given text to a training/known class (topic) or reject/detect it as belonging to some unknown classes~\cite{fei2016breaking,shu2017doc,lin2019deep,zheng2020out}; and \textbf{(2)} \textbf{Semantic novelty detection}, which determines whether a given text represents a semantically/contextually novel phenomenon. For example,~\citet{ma-etal-2021-semantic} studied a semantic novelty detection problem - detecting semantically novel scene descriptions (e.g., ``\textit{A person walks a \underline{chicken} in the park}'' is a novel scene, whereas ``\textit{A person walks a \underline{dog} in the park}'' is normal one). This task is more fine-grained and requires factual reasoning over text as compared to that of (1), which has been studied extensively. (2) has only been introduced recently  and is the focus of this paper.

\begin{figure}[t!]
\centering
\scalebox{0.77}{
\begin{tabular}{cp{6.8cm}|l}
$d_1$~: & ``\textit{\underline{The Big Bang Theory} is an American television sitcom, filmed in front of a live audience, stars \underline{Johnny Galecki} et al.}" & \textbf{normal}           \\    
$d_2$~: &  ``\textit{\underline{Elon Musk}'s performance as a dishwasher in a restaurant in season 9, episode 9 of the \underline{The Big Bang Theory}  is quite interesting to the his fans.}" & \textbf{novel} \\ 
\end{tabular}}
\caption{Examples of semantic novelty detection in factual texts involving named entities.}
\label{fig:problem_exp}
\end{figure}

This paper proposes a new semantic novelty detection task: given a factual text $d$ containing two \textit{named entities}\footnote{Named Entity definition: \url{https://en.wikipedia.org/wiki/Named_entity}}, we want to classify whether $d$ represents a semantically novel fact or a normal one \textit{with respect to} the entity pair. For example, consider the text $d_1$ and an entity pair underlined in $d_1$ in Figure~\ref{fig:problem_exp}. $d_1$ represents a \textit{normal} fact as it is natural for an actor (\textit{Johnny Galecki}) to act in a sitcom or TV show (\textit{The Big Bang Theory}). However, $d_2$ in Figure~\ref{fig:problem_exp} depicts a novel fact with respect to the underlined entity pair because a CEO of a technology company (\textit{Elon Musk}) acting in a sitcom (\textit{The Big Bang Theory}) is quite surprising and novel.

Factual text appears in diverse media sources, such as news articles, blog posts, reviews etc. Detecting semantically novel facts involving popular real-world (named) entities has many applications because anything novel is always of interest and can trigger readers' curiosity. For example, a  mobile newsfeed application can increase user engagement by recommending novel news/facts of named entities and promoting news articles with novel facts. Although novelty is subjective and personal, there exist some novel facts that the majority of people agree. In this work, we restrict our study to this consensus-view of semantic novelty and leave the personalized novelty for future work.

Solving the proposed task requires \textit{joint} fine-grained reasoning over (1) the \textit{relationship} between the pair of entities in the textual context and (2) the \textit{background knowledge} of the entities. For example, considering $d_2$ in Figure~\ref{fig:problem_exp}, we first need to detect that the entity pair (``\textit{Elon Musk}'', ``\textit{The Big Bang Theory}'') in $d_2$ has the ``\textit{cast-member}'' relation and then, leverage the interaction of the relation with the background knowledge of the entities (i.e., ``\textit{Elon Musk}'' is a \textit{tech entrepreneur} and ``\textit{The Big Bang Theory}'' is a \textit{TV show}) to infer the semantic novelty (because, a tech entrepreneur does not normally act in a TV show). We utilize the external Knowledge Repository (KR) - WikiData ~\cite{vrandevcic2014wikidata} to extract the named entity's background knowledge, which is a list of property-value pairs. For example, \textit{Elon Musk}'s background knowledge contains property-value pairs: [(a) (\texttt{occupation}, \textit{entrepreneur}), (b) (\texttt{gender}, \textit{male}), (c) (\texttt{field of work}, \textit{tech entrepreneurship}) ....]. However, not all property-value pairs is useful for inference (e.g., only (a) and (c) are useful for $d_2$ in Figure~\ref{fig:problem_exp}). Thus, a solution for automatic selection of the \textit{useful} property-value pairs is needed (see Sec.~\ref{sec:proposed-model}). {\color{black}In fact, the \textit{useful} property-value pairs provide a reason or \textit{characterization} for the novelty. }

\textbf{Problem Definition}: Given (1) a set of training factual text $\mathcal{D}_{tr}$ = \{$d_1, d_2, .....d_n$\}, with each $d_i \in \mathcal{D}_{tr}$ labeled as normal (\textit{NORMAL} class) with respect to a pair of entities $(e_1^i, e_2^i)$ appeared in $d_i$, and (2) a knowledge base (KB) $\mathcal{K}$ containing the background knowledge (property-value pairs) of a set of entities that is a superset of the entities appeared in $\mathcal{D}_{tr}$, our goal is to build a model $\mathcal{F}$ to score the semantic novelty of a test factual text $d^{\prime}$ having a pair of entities ($e_1^{'}$, $e_2^{'}$) with respect to $\mathcal{D}_{tr}$, $\mathcal{K}$, and pair ($e_1^{'}$, $e_2^{'}$), i.e., classifying $d^{\prime}$ into one of the classes \{\textit{NORMAL}, \textit{NOVEL}\}. As  $\mathcal{F}$ is built with only the ``\textit{NORMAL}'' data, the task is an \textit{one-class classification problem}.

This task is different from the semantic novelty detection task in~\cite{ma-etal-2021-semantic} in two main aspects: (1) Our task demands semantic reasoning over named entities which do not have sufficient semantic information in their textual (or surface) form in $d$. Rich background knowledge of the entities is needed to detect novelty. The task in \cite{ma-etal-2021-semantic} does not require any of such entity background knowledge. (2) \cite{ma-etal-2021-semantic} do semantic reasoning for relations (between objects), based on a fixed/closed set of verbs. However, in our work, the relations between entities may be expressed in any surface forms and/or even implicitly (e.g., the relation ``\textit{cast-member}" between the underlined entities is expressed implicitly in $d_2$). \citet{ma-etal-2021-semantic} cannot handle such cases.

To solve the task, we propose a new model, called PAT-SND (\textit{\textbf{P}roperty \textbf{AT}tention network for \textbf{S}emantic \textbf{N}ovelty \textbf{D}etection}) to detect novel factual text. Additionally, PAT-SND also provides the characterization (or reason) for the novelty (unlike \citet{ma-etal-2021-semantic}). PAT-SND first employs an existing relation classification technique to identify the relation between the entity pair. The identified relation is then used in a novel \textit{relation-aware} \textbf{P}roperty \textbf{AT}tention Network (PAT) module that leverages the attention mechanism to select the useful background knowledge from the KB $\mathcal{K}$ to perform semantic reasoning for novelty detection. The learned attention knowledge in PAT is also used to provide the characterization for the novelty (see Sec.~\ref{sec:proposed-model}). 

PAT-SND is evaluated using our \textbf{newly created} NFTD (\textbf{N}ovel \textbf{F}actual \textbf{T}ext \textbf{D}etection) dataset. We leverage a distant supervision technique~\cite{mintz2009distant} with the Wikipedia\footnote{\tiny \url{https://en.wikipedia.org/wiki/Main_Page}} as the corpus and Wikidata as the KR to build a large training dataset. 
Evaluation results show that PAT-SND outperforms the 10 latest novelty detection baselines by very large margins. 

Our main contributions are as follows:
\begin{enumerate}
	\item We propose a new semantic novelty detection task for factual text involving named entities.
	\item We propose an effective technique called PAT-SND to solve the proposed task.
	\item The proposed technique also provides the characterization of novelty based on the attention knowledge in the PAT-SND model.
	\item A new dataset called NFTD is created for the proposed task as no suitable data is available. The dataset can be used as a benchmark by the NLP community.
\end{enumerate}

\section{Related Work}
\label{sec:related_work}

Novelty or anomaly detection has been studied extensively over the years. Early representative works include \textit{one-class SVM} (OCSVM)~\cite{Bernhard2001,manevitz2001one}, \textit{Support Vector Data Description} (SVDD)~\cite{Tax2004support} and hybrid approaches \cite{Erfani2016high,Ruff2018deep} that learn features using deep learning and then apply OCSVM or SVDD to build one-class classifiers. More recent deep learning approaches are based on auto-encoders~\cite{You201708,Abati2019latent,Chalapathy2019deep}, GAN~\cite{Perera2019ocgan,zheng2019one}, neural density estimation~\cite{Wang2019multi}, multiple hypothesis prediction~\cite{Nguyen2019icml}, robust mean estimation~\cite{Dong2019quantum} and regularization~\cite{hu2020hrn}. \citet{Chalapathy2019deep,pang2020deep} provides a detailed survey. Our PAT-SND is based on an attention network and data augmentation technique.

Novelty detection has also been studied in out-of-distribution (OOD) detection or open-set recognition~\cite{Liang2018enhancing,Shu2018unseen,Erfani2017from,xu2019open}. However, these methods work in the multi-class setting. Ours is an one-class classification problem. There are also works on topical novelty detection~\cite{dasgupta2016automatic, ghosal2018novelty,nandi2020quest,jo2020cnn,li2005novelty,zhang2009combining}. They differ from ours as we focus on \textit{fine-grained} semantic novelty detection.

Our work is also related to \emph{Semantic plausibility} (SPL) that studies the problem of whether an event is plausible or not \cite{porada2019can,wang2018modeling,keller2003using,zhang2017ordinal,sap2019atomic} and \emph{selectional preference} (SPR) that deals with the ``typicality" of an event \cite{resnik1996selectional,clark2001class,erk2010exemplar,bergsma2008discriminative,ritter2010latent,seaghdha2010latent,van2009non,van2014neural,dasigi2014modeling,tilk2016event}. These works differ from ours as {\color{black}(1) conceptually, SPL and SPR are related but different from novelty, (2) Our task demands the use of background knowledge in the named entities for semantic reasoning. However SPL and SPR only perform reasoning on the surface form of objects in the text, and (3) they use fully labeled data~\cite{dasigi2014modeling} while we have only normal (one-class) data in training.}

Commonsense reasoning is remotely related to our work. Existing works have built multi-choice commonsense reasoners~\cite{zellers2018swag,zellers2019hellaswag}, studied the commonsense knowledge contained in language models~\cite{davison2019commonsense,trinh2019do,trinh2018simple} and knowledge graph~\cite{bosselut2019comet}, and constructed new datasets for better evaluation~\cite{wang2020semeval}. Several researchers also investigated physical commonsense reasoning~\cite{bagherinezhad2016elephants,forbes2017verb,wang2017distributional,bisk2020piqa} and affordance of entities~\cite{forbes2019neural}. They do not perform novelty detection.

Trivia fact mining~\cite{merzbacher2002automatic,Ganguly2014AutomaticPO,Gamon2014PredictingIT,prakash2015did,Fatma2017TheUS,mahesh2smart,tsurel2017fun,niina2018trivia,korn2019automatically,kwon2020hierarchical} is also relevant, but it is mainly about interestingness. Some trivia facts are interesting because they are rare, \textit{but not necessarily novel}. Existing papers use labeled training data for learning, or rely on Wikipedia structure to retrieve interesting facts using information retrieval methods~\cite{tsurel2017fun,kwon2020hierarchical}. We have only normal data but no novel data.

Our proposed model is based on an attention network. Related NLP works using attention techniques include \cite{Huang2019SyntaxAwareAL,ma2020entity,Guo2019AttentionGG,wang2020relational,veyseh2020improving,9177070}. But they solve different problems, such as sentiment analysis and argument mining and are not about novelty detection. Their approaches also differ from ours. 

\section{Dataset Collection and Annotation}
\label{sec:dataset-collection-annotation}

To build a large factual text dataset annotated with named entities, we leverage the distant supervision technique in~\citet{mintz2009distant}. We {create our training and test datasets}, using Wikipedia as the corpus and Wikidata~\cite{vrandevcic2014wikidata} as the external Knowledge Repository (KR). 

We choose Wikidata as KR for extracting background information of the entities, because the good community collaboration and contribution of Wikidata makes it a high-quality KR compared to other KRs~\cite{farber2015comparative}. Wikidata encodes real-world knowledge in the form of triples: ($e_1$, $r$, $e_2$), which means entity $e_1$ and entity $e_2$ have a relation $r$.  For instance, (The Big Bang Theory, Cast-Member, Johnny Galecki).

The named entities in the Wikipedia corpus are linked to the Wikidata. We can find unambiguous mappings between entity mentions in the text and Wikidata entities. \textcolor{black}{For example: In the \textit{Wikipedia Source}: ``[[The Big Bang Theory]] is an American television sitcom, filmed in front of a live audience, stars [[Johnny Galecki]] et al.'', the named entities in bracket [[.]] have an unique one-to-one mapping to the entities in Wikidata.}

\paragraph{Training dataset preparation.} \textcolor{black}{The distant supervision technique can be briefly described as follows:} For a piece of text $d$ from Wikipedia involving $e_1$ and $e_2$ (with hyperlink uniquely mapping to Wikidata entities), if there is a triple $(e_1, r, e_2)$ in the KR, we assume that the textual information in $d$ expresses the relation $r$ between $e_1$ and $e_2$. In this case, we automatically annotate $(e_1, r, e_2, d)$ as a distantly supervised instance and add it to our training dataset. For entity pairs $(e_1, e_2)$ with more than one relation, we discard them because they bring ambiguity in our dataset. Due to the budgetary constraints, we can not evaluate on all relations in the Wikidata. 

We create our training data related to 20 human related relations. The details of these 20 relations are in Appendix Sec.~\ref{appendix-sec::dataset-details}. With distant supervision, we allow noise to exist in the training dataset because this process requires no human annotation, and scales up the learning of more relations. We split \textcolor{black}{the whole dataset created via distant supervision} into two parts: \textit{train set} and the \textit{test set pool}, making sure that there is no overlapping in either text or entity pairs between these two parts. This \textit{test set pool} is used for test dataset preparation.

\begin{table}[t!]
\centering
\caption{NFTD dataset statistics. NR (NV) denotes the NORMAL (NOVEL) class. ``text length'' is \# of words.}
\scalebox{0.7}{
\begin{tabular}{c|c|c}
\hline
 & \textbf{Training} & \textbf{Test} \\ 
\hline
\# instances (factual text) & 251,619 (NR) & 1000 (NR), 1000 (NV) \\ 
Avg. text length & 41.35 & 26.02 \\
\hline
\end{tabular}
}
\label{table:data_stat}
\end{table}

\paragraph{Test dataset preparation.} While training dataset may contain noise, test data needs to be manually annotated and checked for a fair evaluation. We invited five graduate students with advanced level of English as crowd workers. We randomly split the \textit{test set pool} into two parts: \textit{normal test data pool-1} and \textit{normal test data pool-2}. 

\textbf{Normal test data.} We assume that the fact descriptions in Wikipedia are all normal facts. So for normal test data, we sample instances from the \textit{normal test data pool-1} and assign them to annotators to identify the correct instances. Each instance is a tuple $(e_1, r, e_2, d)$. The annotators are asked to check whether or not the sentence $d$ with the entity pair $(e_1, e_2)$ semantically expresses the relation $r$. If yes, this instance is added to our normal test dataset. After an instance is collected, we ensure that it is verified by another annotator. If there is a disagreement, we make sure it is discussed and resolved between the two annotators. Following this procedure, we annotate 50 normal instances for each relation. 

\textbf{Novel test data.} We divide the whole task into 20 subtasks and evenly assign them to the annotators. For each subtask, the goal is to generate 50 novel tuples $(e_1, r, e_2, d)$ for each relation. Instead of asking annotators to create novel instances from scratch, we sample some instances from \textit{normal test data pool-2} to inspire annotators. They are asked to change the property-value pairs of entities and the text $d$, or even write from scratch if they come up with interesting ideas. 

After the first round annotation, we get 50 novel instances for each of the 20 relations. Then, the annotations are shown to the other four annotators to label them as normal or novel. \textcolor{black}{We use the majority voted label as the final label of these instances. We use Fleiss’ Kappa~\cite{fleiss1973equivalence} to calculate the inter-rater reliability. The Fleiss’ Kappa score is 0.91, interpreted as high agreement, which means our test data reflects the consensus-view of semantic novelty.} At the end, we collect 50 normal and 50 novel instances for each of the 20 relations. Table~\ref{table:data_stat} shows the NFTD dataset statistics. 

The details of the data annotation guideline is in Appendix Sec.~\ref{appendix-sec:annoation-guideline}.

\paragraph{Building Entity Background KB ($\mathcal{K}$).}
\label{sec:background-kb}

We use the knowledge repository (KR), Wikidata, to build the entity background KB $\mathcal{K}$. KR is represented as: $KR=(\mathcal{E},\mathcal{R},\mathcal{T})$, where $\mathcal{E}$ denotes a set of entities, $\mathcal{R}$ is a set of relations/edges, and $\mathcal{T} \subseteq \mathcal{E} \times \mathcal{R} \times \mathcal{E} $ is the set of all triples. For each entity $e$ in $\mathcal{E}$, we obtain the list of property-value pairs as $e$'s background knowledge to build $\mathcal{K}$ as follows.

We first collect all triples from KR involving $e$ and then extract the relation and the other entity from each triple to form a property-value pair with the relation as a property and the other entity as the value of the property. For example, considering $e$ = ``\textit{Elon Mask}'' and a triple (``\textit{Elon Mask}'', ``\textit{occupation}'', ``\textit{entrepreneur}'') in KR, the extracted property-value pair for $e$ would be  (\texttt{occupation}, \textit{entrepreneur}).

Let $\mathcal{P}$ be the complete property set in the background KB $\mathcal{K}$. We assume that each $e_i$ in the training data is in the $\mathcal{K}$. However, $e_i$ in the test data can be a new entity (i.e., it does not appear in the training data), as long as the background knowledge of the entity is available to our model (where, the property-value pairs are either retrieved from the KR or provided by the human annotator during the test data annotation process and included in $\mathcal{K}$).

\section{Proposed Approach}
\label{sec:proposed-model}

Our proposed PAT-SND model works in two steps: (1) \textit{Entity Relation Classification}, and (2) \textit{Triple Semantic Novelty Scoring} (SNS). Given a factual text $d$ containing a pair of entities $(e_1, e_2)$, PAT-SND first identifies the relation $\hat{r}$ between $(e_1, e_2)$ in $d$ in step (1) [Sec.~\ref{sec:rc-classification}]. Next, the background knowledge of the entities $e_1$ and $e_2$ retrieved from the KB $\mathcal{K}$ together with  the predicted relation $\hat{r}$ are fed to the SNS module to score the semantic novelty of $d$ with respect to $(e_1, e_2)$ and $\mathcal{K}$ in step 2 [Sec.~\ref{sec:sns}]. As our training data $\mathcal{D}_{tr}$ consists of only NORMAL class examples (as discussed in Sec.~\ref{sec:intro}), it's not possible to train SNS solely with $\mathcal{D}_{tr}$. Thus, we  propose a \textit{KB-based Contrastive Data Generator} (CDG) to generate pseudo-novel examples. The SNS module is then trained with both NORMAL class examples in $\mathcal{D}_{tr}$ as well as the generated pseudo-novel examples in a supervised learning manner. We will discuss more about it in Sec.~\ref{sec:training}.

\subsection{Entity Pair Relation Classification}
\label{sec:rc-classification}
Given a factual text $d$ having entity pair $(e_1, e_2)$, we build a model to identify the relation $\hat{r}$ between $(e_1$ and $e_2)$ in $d$. For this purpose, we utilize a BERT-based Relation Classification model~\cite{wu2019enriching}, that incorporates entity position information into a  pre-trained language model for relation classification. Next, we combine the identified relation $\hat{r}$ with the entity pair to produce a triple ($e_1$, $\hat{r}$, $e_2$) which serves as input to the SNS (in Sec.~\ref{sec:sns}). 

During training process, the relation classification model is trained using $\mathcal{D}_{tr}$, where each $d_i \in \mathcal{D}_{tr}$ is labelled with \textit{true relation label} $r$ between the entity pair through the distant supervision technique.

\subsection{Triple Semantic Novelty Scoring (SNS)}
\label{sec:sns}

Let $B_1 = \{(p_i^1, v_i^1) | 1 \leq i \leq l \}$ and $B_2 = \{(p_i^2, v_i^2) | 1 \leq i \leq m \}$ be the background knowledge obtained for $e_1$ and $e_2$ respectively from KB $\mathcal{K}$ (See Sec.~\ref{sec:background-kb}). The SNS module utilizes $B_1$, $B_2$ and relation $\hat{r}$ as inputs to score the novelty of the input text $d$. In this process, SNS employs a relation-aware attention mechanism over $B_1$ and $B_2$ to select the useful knowledge, which is motivated as follows.

\begin{figure}[t!]
\centering
  \includegraphics[scale=0.52]{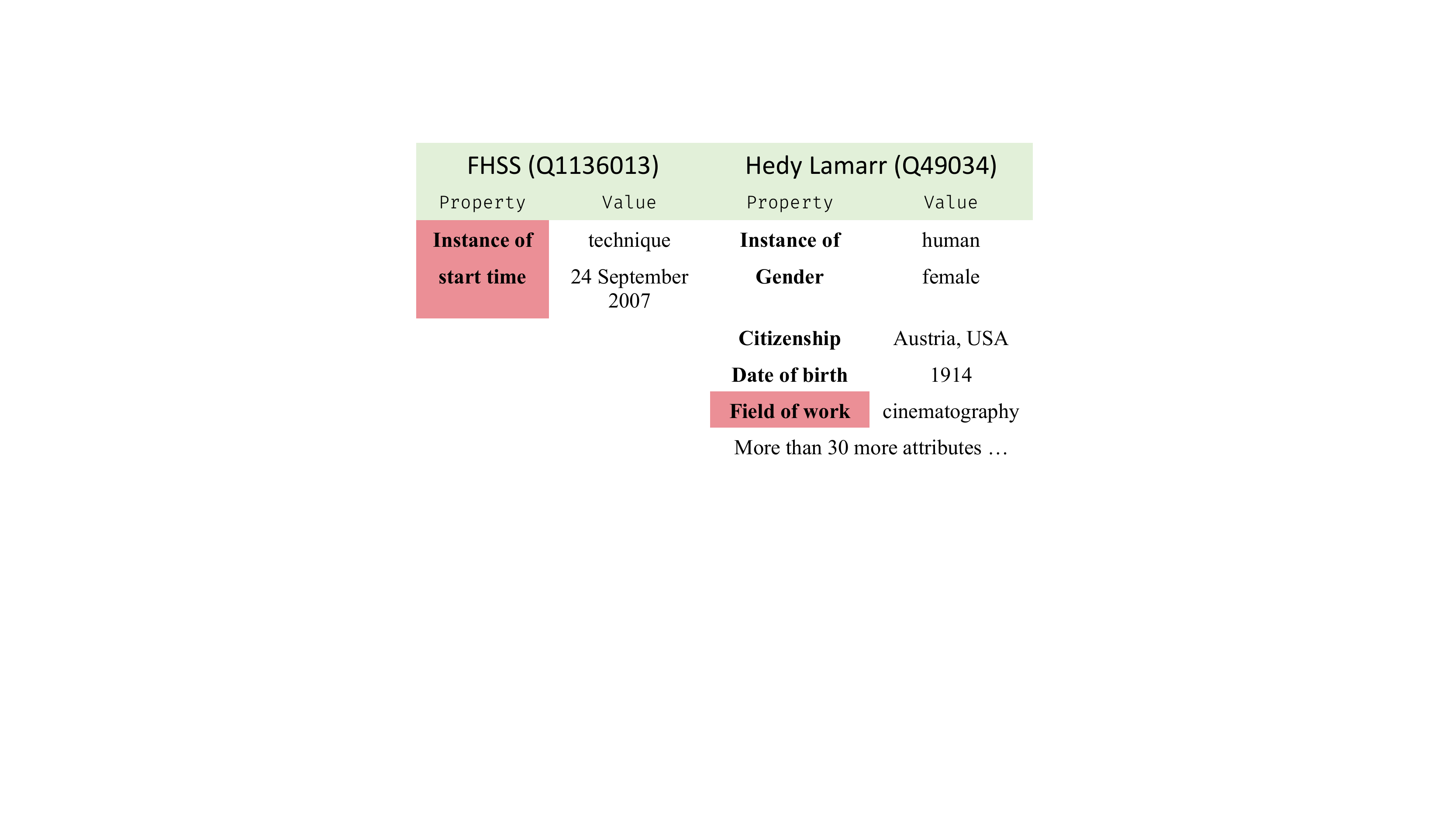}
  \caption{Illustration of two entities' property and value pairs in the KB $\mathcal{K}$. The properties marked in red are useful or important for detecting semantic novelty of the example $d_2$ in Figure 1.}
  \label{fig:property-value-pairs}
\end{figure}

Leveraging all property-value pairs in $B_1$ and $B_2$ may not be helpful to detect the novelty of the text $d$. For example, as shown in Figure~\ref{fig:property-value-pairs}, considering the entity ``\textit{Elon Mask}", the property-value pair (\texttt{occupation}, \textit{entrepreneur}) is useful to score the novelty of $d_2$ in Figure 1, whereas (\texttt{gender}, \textit{male}) is not useful at all. Thus, the model needs to have the ability to focus on important information and filter out noises in $B_1$ and $B_2$. Such knowledge selection process is relation dependent, as for different relations, different property-value pairs would be useful for novelty detection. 

To enable automated knowledge selection, SNS is built using a key component called \underline{P}roperty \underline{At}tention Network (PAT) that utilizes the semantics of the relation $\hat{r}$ to attend over $B_1$ and $B_2$ for inference. As the attention mechanism needs to be relation-specific, we build one PAT module for each relation. So, for detecting novelty of a test text $d'$, SNS fires the PAT learned for relation $\hat{r}$, identified from $d'$ using the Relation Classifier (in Sec.~\ref{sec:rc-classification}).  

\paragraph{Property Attention Network (PAT).} PAT takes a list of property-value pairs $\{(p_i, v_i) | 1 \leq i \leq N\}$ and a relation $r$ as input and outputs a weighted value vector $\bm h^{out}$ to be used for inference. $p_i$ and the corresponding $v_i$ are fed to PAT as feature vectors $\bm p_i$, $\bm v_i$ respectively, together with $r$ (to invoke the relation-specific module). We employ BERT~\cite{Devlin2019bert} to learn the embedding representation of $p_i$ , $v_i$ and use them as corresponding feature vectors. For example, the property ``\textit{instance of}'' is encoded as $\langle$[CLS], instance, of, [SEP]$\rangle$ using WordPiece Tokenizer and fed into BERT and embedding corresponding to token [CLS] in the output layer of BERT is used as the feature vector of the property.

In PAT, the \{$\bm p_i$\}$_{i=1}^N$ are fed one by one through a relation-specific linear layer, and a \textit{relu} non-linearity function and a softmax function are used to obtain the attention weights \{$\bm \alpha_{ir}$\}$_{i=1}^N$ over \{$\bm p_i$\}$_{i=1}^N$ with respect to $r$. Next, the weights are used to weigh the corresponding \{$\bm v_i$\}$_{i=1}^N$ to obtain $\bm h^{out}$. The processing for a given $r$ is summarized:

\begin{equation}
  \scalebox{0.8}{
      $\begin{split}
          \bm g_{ir}^{k} = relu(\bm p_i ~\bm W_{r}^{k} + \bm b_{r}^{k})  \\
          \bm \alpha_{ir}^{k} = \frac{exp(\bm g_{ir}^{k})}{\sum_{i=1}^{N}exp(\bm g_{ir}^{k})} \\
          \bm h^{out} = \sum_{i=1}^{N} (\frac{1}{K}\sum_{k=1}^{K} \bm \alpha_{ir}^{k}) ~\bm v_i^{k}
      \end{split}$
      }
      \label{eq:PAT_layer}
\end{equation}

where $K$ is the total number of attention heads and $\bm W_{r}^{k}$, $\bm b_{r}^k$ are relation-specific weight and bias for the $k$-th attention head. $\bm \alpha_{ir}^k$ is the $k$-th attention weight between $r$ and $p_i$. Overall, the processing of inputs in PAT is denoted as \scalebox{0.8}{$h^{out} = PAT(\boldsymbol{P}, \boldsymbol{V}, r; \Theta_r)$}, where \scalebox{0.8}{$\boldsymbol{P} = [\bm p_1, \bm p_2, ..., \bm p_N] \in \mathbb{R}^{N \times {F}}$} is the property matrix, \scalebox{0.8}{$\boldsymbol{V} = [\bm v_1, \bm v_2, ..., \bm v_N] \in \mathbb{R}^{N \times {F}}$}, is the value matrix and $\Theta_r$ is the trainable parameters for relation $r$.

\textbf{Triple Novelty scoring.} Given the inputs $B_1$, $B_2$ and relation $\hat{r}$, we obtain the property and value matrices $\boldsymbol{P_1}$, $\boldsymbol{V_1}$ from $B_1$ and $\boldsymbol{P_2}$, $\boldsymbol{V_2}$ from $B_2$ and feed them to PAT for relation $\hat{r}$ as follows:

\begin{equation}
\small
    \begin{split}
        h^{out}_1 &= PAT(\boldsymbol{P_1}, \boldsymbol{V_1}, \hat{r}; \Theta_{\hat{r}}) \\
        h^{out}_2 &= PAT(\boldsymbol{P_2}, \boldsymbol{V_2}, \hat{r}; \Theta_{\hat{r}}) \\
        h^{out}_{\hat{\tau}} &= [h^{out}_1 ; h^{out}_2] \\
    \end{split}
\end{equation}

Next, a relation-specific feed-forward layer is used to project $h^{out}_{\hat{\tau}}$ into a semantic novelty score as $S(\hat{\tau}) = (h^{out}_{\hat{\tau}}~\bm W_{\hat{r}} + \bm b_{\hat{r}})$, where $\hat{\tau}$ denotes the triple ($e_1$, $\hat{r}$, $e_2$). Following the existing one-class classification literature~\cite{Chalapathy2019deep,pang2020deep}, we do not use a threshold to further produce a classification label, instead use $S(\hat{\tau})$ directly in our experiments (Sec.~\ref{sec:experiments}).

\subsection{Training}
\label{sec:training}

Let $\mathcal{T}_{tr}$ be the set of all triples (labelled as NORMAL class) extracted from the examples in $\mathcal{D}_{tr}$. To train SNS, we use KB $\mathcal{K}$ to help generate contrastive examples (triples) by corrupting the triples in $\mathcal{T}_{tr}$, as discussed below. These contrastive examples serve as the pseudo-novel data and enable the supervised learning of the SNS.

\paragraph{KB-based Contrastive Data Generator.} Given a triple $\tau_i \in \mathcal{T}_{tr}$, the generator $G_{contrastive}(\tau_i)$ randomly samples an entity $e^{'}$ from KB $\mathcal{K}$ to replace either $e_1$ or $e_2$ in $\tau_{i}$. After corruption, $\tau_i^{'}$ is formed from $\tau_{i}$, where $\tau_i^{'} = ({e^{'}, r, e_2})$ or $\tau_i^{'} = (e_1, r, e^{'})$. For example, given $\tau_1$ = (\textit{The Big Bang Theory}, \textit{cast-member}, \textit{Johnny Galecki}) as a NORMAL triple in $\mathcal{T}_{tr}$, a pseudo-novel triple generated by $G_{contrastive}(\tau_1)$ would be $\tau_1^{'}$ = (\textit{The Big Bang Theory}, \textit{cast-member}, \underline{\textit{Warren Buffett}}). During the training of SNS, we dynamically generate one pseudo-novel triple for each NORMAL triple in $\mathcal{T}_{tr}$ in every training epoch. 

\textbf{Learning.} PAT-SND is trained end-to-end by minimizing a max-margin ranking objective as,
\begin{equation}
\small
    \mathcal{L} = \sum_{\tau \in \mathcal{T}_{tr}} \sum_{\tau^{\prime} \in {\mathcal{T}^{\prime}_{tr}}} max\{S(\tau^{\prime}) - S(\tau) + 1, 0\}
\label{eq:loss}
\end{equation}

where, ${\mathcal{T}^{\prime}_{tr}}$ is the set of pseudo-novel triples generated from $\mathcal{T}_{tr}$. $\mathcal{L}$ encourages the score $S(\tau)$ of the NORMAL triple $\tau$ to be higher than $S(\tau^{\prime})$ of a pseudo-novel triple $\tau^{\prime}$.

\section{Experiments}
\label{sec:experiments}

\subsection{Experiment Setup}
The details of the dataset annotation and statistics have been discussed in Sec.~\ref{sec:dataset-collection-annotation}. All the results reported in this section are the averages of five runs with different random seeds. The code and the dataset are released\footnote{The Github for released code and the annotated data: \url{https://github.com/NianzuMa/PAT-SND}}.

\paragraph{Evaluation Metrics.} Since our task is an one-class classification task, we follow the existing one-class classification literature~\cite{Chalapathy2019deep,pang2020deep} and use AUC (Area Under the ROC curve) as the evaluation metric.

\paragraph{Baselines.}~Since the proposed task is new, we are not aware of any existing model that can be directly applied to our task. We converted two types of existing methods to be used as Semantic Novelty Scorers (SNS) for our task: (i) \textbf{language models (LMs)} , and (ii) \textbf{traditional and deep learning based one-class classifiers}. Note that, the GAT-MA in~\cite{ma-etal-2021-semantic} model cannot be used as a baseline because the model needs verbs expressed explicitly in text for novelty scoring. However, in our case, the relation in the factual text may be implicitly expressed in various surface forms, which makes GAT-MA inapplicable to our task.

(i) \textbf{LM-based SNS.} We train LMs on our training text data, which are all normal factual text. When the LMs are trained to minimize the perplexity of text, it maximizes the probability of the words appearing in the text context. The trained models thus capture the semantic meaning of the words and the text. If something unexpected appears in the context, the model has the ability to detect the novelty. The trained language models are used first to output the probability of each word in the text, and then we calculate the sentence probability based on these word probability scores. Following~\cite{ma-etal-2021-semantic}, we use (a) arithmetic mean, (b) geometric mean, (c) harmonic mean, and (d) multiplication of all word probabilities. We find that harmonic mean gives the best results. Among language models, we adopt \textbf{N-gram}, the bag of words LM, $N \in \{1, 2, 3, 4, 5\}$ ($N=5$ gives the best result), \textbf{BERT}~\cite{Devlin2019bert}, \textbf{GPT-2}~\cite{radford2019language} as our LM-based SNS and show the results in Table~\ref{table:model-results}.

(ii) \textbf{One-class Classifier based SNS.} One-class classification methods~\cite{perera2021one} aim to identify instances of a specific class amongst all instances, by primarily learning from a \textit{training set containing only the instances of that class}. There is a considerable amount of research that has been done in the computer vision, machine learning, and biometrics communities. While most of them are designed for image data, we convert the models to SNSs by modifying the feature encoder parts of the models. Here are the classical statistical and recent deep learning-based one-class classifiers: 

(1) \textbf{OCSVM}~\cite{Bernhard2001}: the classic one-class SVM classifier. (2) \textbf{iForest}~\cite{Liu2008isolation}: an ensemble method using random unsupervised trees. (3) \textbf{VAE}~\cite{Kingma2014auto}: a variational auto-encoder used as one-class classifier. (4) \textbf{OCGAN}\cite{Perera2019ocgan}: a popular one-class novelty detection model based on GAN. (5) \textbf{DSVDD} (Deep SVDD)~\cite{Ruff2018deep}: a deep learning implementation of the one-class classifier SVDD~\cite{Tax2004support}. (6) \textbf{ICS}~\cite{Schlachter2019deep}: an one-class classifier trained using the training data split into two parts: typical and atypical. (7) \textbf{HRN}~\cite{hu2020hrn}: a recent model based on a holistic regularization method. We do not compare with other models that require image transformation such as CSI~\cite{tack2020csi}. Out-of-distribution (OOD) detection methods are not applicable to our task since they typically need multiple classes to train the model.

The details of experiment settings are provided in the Appendix Sec.~\ref{appendix-sec:gat-implementation}.

\subsection{Novelty Detection Results and Analysis}
\label{sec:result-and-analysis}

\begin{table*}[t!]
\centering
\caption{Comparison of baselines and our proposed model (based on AUC score). Each result in the table is the average of 5 runs with different seeds ($\pm$ standard deviation).}
\scalebox{0.80}{
\begin{tabular}{ccc|ccccccc|c}
\hline
  \multicolumn{3}{c|}{Language model based model} & \multicolumn{7}{c|}{General One-class classifier} & Proposed \\
 \hline
 Ngram & BERT & GPT-2 & OCSVM & iForest & VAE & DSVDD & ICS & OCGAN & HRN & \textbf{PAT-SND} \\ 
 \hline
 50.02\stdv0.0 & 60.12\stdv0.0 & 58.13\stdv0.0 & 50.63\stdv0.0 & 44.16\stdv1.3 & 47.94\stdv0.3 & 51.00\stdv0.5 & 53.98\stdv0.5 & 52.10\stdv0.0 & 55.53\stdv1.3 & \textbf{90.37\stdv0.5} \\
 \hline
\end{tabular}
}
\label{table:model-results}
\end{table*}

\begin{figure*}[ht!]
\centering
  \includegraphics[scale=0.6]{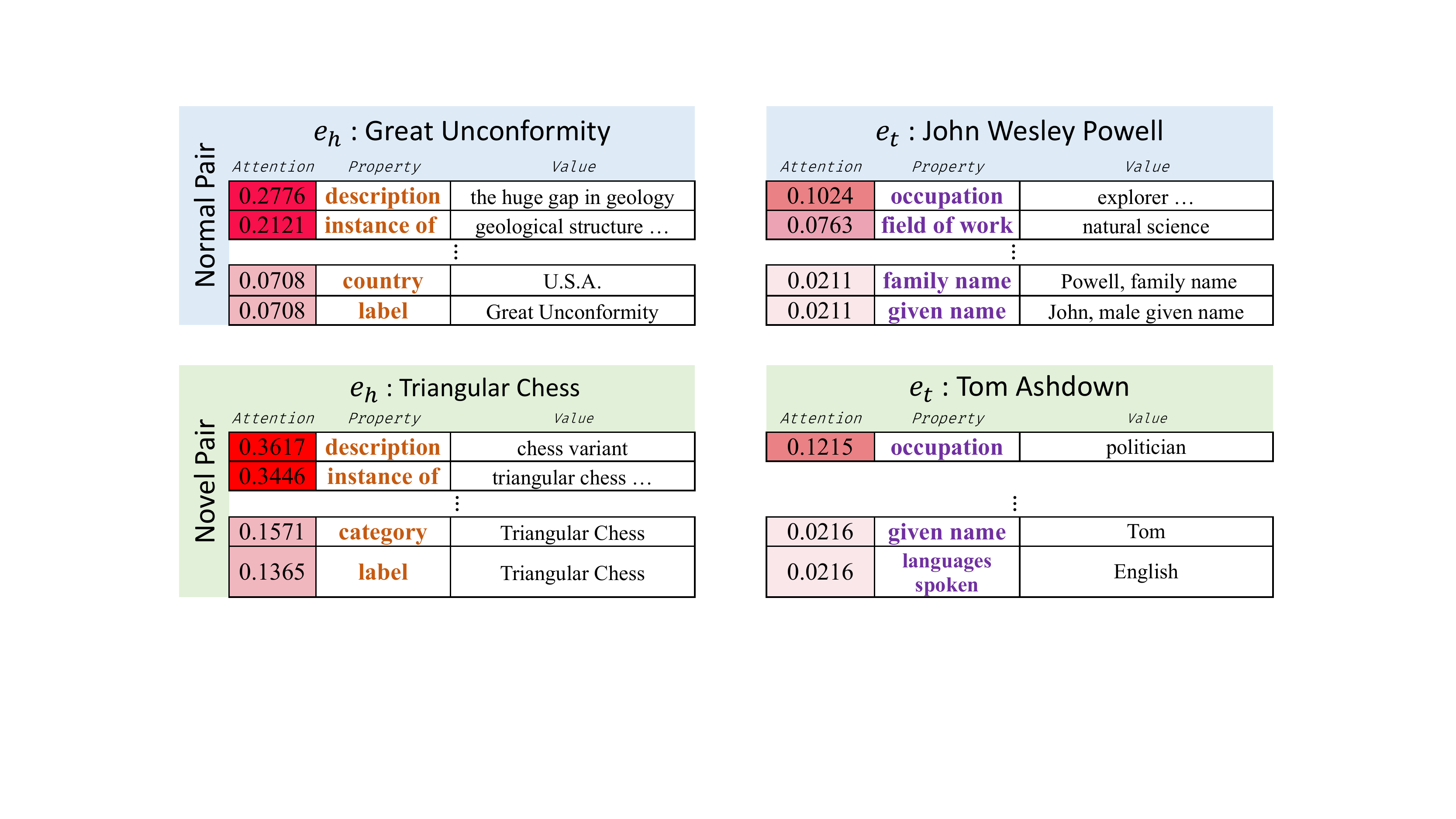}
  \caption{PAT-SND attention illustration for relation ``discoverer/inventor'' on a normal and a novel entity pairs.}
  \label{fig:entity-pair-attention}
\end{figure*}

\paragraph{Model Comparison and Discussion.} We show the results of all baselines and our proposed model PAT-SND in Table~\ref{table:model-results}. Here are the conclusions we can drawn from the results:

\textbf{(1)} All LM-based SNSs perform poorly on our factual text novelty detection task, \textcolor{black}{because although they implicitly learn the syntactic and semantic information of the text, they cannot explicitly do semantic reasoning.} The information in text alone is not enough to distinguish normal and novel factual text. Our task needs the background information (property-value pairs) of named entities to perform semantic reasoning and detect novelty. The language models dealing with sequential data can hardly incorporate  background knowledge of named entities during training.

\textbf{(2)} All one-class classifier based SNSs also perform poorly on our task. To employ the one-class classifiers, we first extract the text embedding using a text encoder and then use the embedding to learn the classifier. The text encoder parameters are frozen during the classifier training. The text embedding is computed by averaging the token embeddings obtained from the last layer of BERT (used as text encoder in our baselines). However, none of these methods are able to incorporate background knowledge of the named entities into the embedding. Thus, they perform poorly on our task.

For our proposed method, the macro F1 score of relation classification (Sec.~\ref{sec:rc-classification}) is 95.12\%. PAT-SND's novelty detection AUC score is 90.37, which is better than the AUC score of all baselines by large margins. \textcolor{black}{We believe the reasons are: 
(1) our model exploits the background knowledge of the two named entities to do semantic reasoning, which is a necessity for our task. (2) the contrastive data augmentation converts our task into a supervised learning problem, enabling our model to be trained to select important relation-specific properties and values to do effective semantic reasoning.}

\subsection{Novelty Characterization}
\label{sub-sec:novelty-characterization}

\textbf{Case Study - PAT-SND attention illustration.} We analyze one normal and one novel factual text here: (1) NORMAL: \textit{``The term \underline{Great Unconformity} is frequently applied to the unconformity observed by \underline{John Wesley Powell} in the Grand Canyon in 1869''}. (2) NOVEL: \textit{``The best known is a chess variant for two players, \underline{Triangular Chess}, invented by \underline{Tom Ashdown} in 1986''}. In Figure~\ref{fig:entity-pair-attention}, we illustrate the property attention from PAT-SND for the normal and novel entity pairs, which represent how  each property contributes to the semantic reasoning with respect to the relation ``discoverer/inventor.'' The property-value pair is ranked in decreasing order of the attention weights. We display the most important and the least important entries for each entity in the entity pair. 

As shown in Figure~\ref{fig:entity-pair-attention}, when the model performs semantic reasoning, the model is trained to inspect whether or not the entity $e_1$'s properties ``\texttt{description}'' and ``\texttt{instance of}'' are matched with the entity $e_2$'s properties ``\texttt{occupation}'' and ``\texttt{field of work}''. These trained attention weights of the model align well with our intuition. For the novel entity pair in Figure~\ref{fig:entity-pair-attention}, the trained model successfully focus on the property ``\texttt{occupation}" with value ``\textit{politician}'' of entity ``Tom Ashdown''; the property ``\texttt{description}" with value ``\textit{chess variant}'' and the property ``\texttt{instance of}'' with value ``\textit{triangular chess}'' of entity ``Triangular Chess''. This attention knowledge implies that ``\textit{Tom Ashdown, who is a politician (occupation), invented a triangular chess}'' is unexpected and thus novel.

\textbf{PAT-GAT as a Normal Knowledge Miner.} As we have discussed in the case study above, the attention weights in the PAT-SND model provide knowledge about the importance of property-value entries across all property-value list in two named entities. Since PAT-SND is trained on both normal and pseudo-novel instances, it can not only detect novelty but also normal instances for each relation. Similar to Figure~\ref{fig:entity-pair-attention}, we demonstrate 2 normal and 2 novel examples for all 20 relations in Appendix~Sec.~\ref{appendix-sec:att-illustration}. After inspecting the normal instances for 20 relations in the dataset, we can quickly summarize the normal knowledge mined by the PAT-SND model in natural language. 

For instance, in Appendix~\ref{appendix-sec:att-illustration} Table~\ref{table:relation-examples-3}, for relation ``cast-member'', PAT-SND model shows that the most important properties for $e_1$ are ``\texttt{description}'', ``\texttt{instances of}'', ``\texttt{genre}'' and the most important properties for $e_2$ are ``\texttt{occupation}'', ``\texttt{description}''. Together with the corresponding values of these properties, we can summarize the normal knowledge as ``\textit{an actor is the cast member of a film (TV series or other similar entities)}''. In the same way, we summarize the normal knowledge in natural language for all 20 relations in Table~\ref{table:common-sense-all-relation} (see Appendix~Sec.~\ref{appendix-sec:common-sense-knowledge}). Because the 20 relations in our experiment are not domain-specific, the normal knowledge presented in Table~\ref{table:common-sense-all-relation} is actually common sense knowledge\footnote{The common sense knowledge in NLP is ``broadly reusable background knowledge that's not specific to a particular subject area... knowledge that you ought to have.''~\cite{pavlus2020common}}.

\textbf{Quantitative Analysis.} As we have discussed above, considering relation - ``discoverer/inventor'', \{``\texttt{description}'', ``\texttt{instance of}''\} is the key property set for entity $e_1$ and \{``\texttt{occupation}'', ``\texttt{field of work}''\} is the key property set for entity $e_2$, when the model performs semantic reasoning through the interaction of these entities for novelty detection. From Sec.~\ref{sec:sns}, we see that the higher the attention weights that the model assigns to the key properties, the more effective the model is in detecting semantic novelty and at the same time, produce more accurate characterization of the novelty. 

To quantitatively analyze the model's performance of novelty characterization, we have sampled 100 novel instances from the test dataset and asked two annotators to independently annotate the key property set for entities $e_1$ and $e_2$. For instance, for the novel entity pair in Figure~\ref{fig:entity-pair-attention}, the key property for the entity $e_1$ is \{``\texttt{description}'', ``\texttt{instance of}''\}, the key property set for entity $e_2$ is \{``\texttt{occupation}''\}. After the annotation, the two annotators compare the annotation of each others and discuss to resolve the conflicts (we observed 10 entities out of the 200 named entities to have such conflicts). 

We then design a \textbf{Novelty Characterization Score} (NCS) as follows: we rank the properties for both $e_1$ and $e_2$ based on the attention score in decreasing order. If one of the key properties appear in the Top-N properties of the entity $e_1$, we give it the score 0.5. We follow the same for entity $e_2$. So for each instance, the full score is 1. We calculate the average of the NCS across all 100 instances for Top-1, Top-2, and Top-3 scores and show the result in Table~\ref{table:characterization-model-results}. Since there is no existing method that is able to perform this task, we compare the result with a random model, in which the property rank lists are randomly shuffled. From Table~\ref{table:characterization-model-results}, we see that PAT-SND model outperforms the  ``Random'' baseline by a very large margin.

\begin{table}[t!]
\centering

\caption{Characterization Performance Comparison of baseline and our proposed model (based on  Novelty Characterization Score)}

\scalebox{0.9}{
\begin{tabular}{c|cccc}
  \hline
  Model & Top-1 & Top-2 & Top-3 \\
  \hline
  PAT-SND & 0.82 &  0.96 & 0.97 \\
  Random & 0.16 & 0.29 & 0.40  \\
  \hline
\end{tabular}
}
\label{table:characterization-model-results}
\end{table}

\section{Conclusion}
This paper proposes a new semantic novelty detection problem - Semantic Novelty Detection in Factual Text Involving Named Entities. A novel attention-based network PAT-SND is proposed to solve the problem. A new dataset NFTD is created and released as a benchmark by the NLP community. Experimental results showed that PAT-SND outperforms 10 baselines by very large margins.

\section{Limitations}

~~~~\textbf{Error Propagation.} The proposed model PAT-SND is structured in a pipeline fashion and processes the input in two steps: (1) relation classification and (2) semantic reasoning on the property-value list of the entity pairs. Since this model is not designed as an end-to-end model, errors from step 1 can propagate to step 2. Designing an end-to-end model to alleviate error propagation is an interesting direction to explore in our future work. 

\textbf{PAT-SND Model's Parameter Size.} In the current PAT-SND model design, for each relation, we train a relation-specific module with an attention technique to perform semantic reasoning. When the number of relations grows, the parameter size of PAT-SND will grow linearly, which is not optimal when the number of relations is large. 

We also noticed that the most important property sets for some relations are similar. It is better that the model takes relation $r$ as input and encourage knowledge (parameter) sharing between similar relations. One way of achieving this is through multi-task learning. Its downside is that whenever a new relation is added, the model needs to be retrained, which is very time-consuming. Another way is through continual learning to incrementally learn each relation in a single neural network. However, it comes with the challenge of dealing with catastrophic forgetting, which often causes degradation in model performance. In our future work, we will address these issues.

\textbf{Closed-World Semantic Reasoning.} For relation classification, our model is limited to the relations already defined in the KR. Although the relation defined in the KR is rich, it is not exhaustive. Our model cannot deal with relations that are not present in the KR. This is an interesting direction to explore in the future as well.  

\section*{Acknowledgments}
{\color{black}The work was supported in part by a DARPA contract HR001120C0023 and three National Science Foundation (NSF) grants (IIS-1910424, IIS-1838770, and CNS-2225427).}

\bibliography{anthology,Rebiber_output}
\bibliographystyle{acl_natbib}

\newpage

\appendix

\section{Dataset Details}
\label{appendix-sec::dataset-details}

Due to budgetary constraints, we can not evaluate all relations in the Wikidata. We limit our training data relation to 20 human-related relations. The details such as the Wikidata relation ids, labels, and descriptions of these 20 relations are shown in Table~\ref{table:relation-info}.

\begin{table*}[t!]
\centering
\caption{20 Human Related Relation Information}
\scalebox{0.8}{
\begin{tabular}{c|C{3cm}|p{10.5cm}}
  \hline
  Relation IDs & Label & Description \\
  \hline
  P6 & head of government & head of the executive power of this town, city, municipality, state, country, or other governmental body \\ 
P39 & position held & subject currently or formerly holds the object position or public office \\ 
P57 & director & director(s) of film, TV-series, stageplay, video game or similar \\ 
P58 & screenwriter & person(s) who wrote the script for subject item \\ 
P61 & discoverer or inventor & subject  who discovered, first described, invented, or developed this discovery or invention \\ 
P84 & architect & person or architectural firm responsible for designing this building \\ 
P86 & composer & person(s) who wrote the music [for lyricist, use "lyrics by" (P676)] \\ 
P161 & cast member & actor in the subject production [use "character role" (P453) and/or "name of the character role" (P4633) as qualifiers] [use "voice actor" (P725) for voice-only role] \\ 
P170 & creator & maker of this creative work or other object (where no more specific property exists). Paintings with unknown painters, use "anonymous" (Q4233718) as value. \\ 
P175 & performer & actor, musician, band or other performer associated with this role or musical work \\ 
P241 & military branch & branch to which this military unit, award, office, or person belongs, e.g. Royal Navy \\ 
P412 & voice type & person's voice type. expected values: soprano, mezzo-soprano, contralto, countertenor, tenor, baritone, bass (and derivatives) \\ 
P413 & position played on team / speciality & position or specialism of a player on a team \\ 
P463 & member of & organization, club or musical group to which the subject belongs. Do not use for membership in ethnic or social groups, nor for holding a position such as a member of parliament (use P39 for that). \\ 
P641 & sport & sport that the subject participates or participated in or is associated with \\ 
P800 & notable work & notable scientific, artistic or literary work, or other work of significance among subject's works \\ 
P991 & successful candidate & person(s) elected after the election \\ 
P1303 & instrument & musical instrument that a person plays or teaches or used in a music occupation \\ 
P1346 & winner & winner of an event or an award; on award items use P166/P1346 on the item for the awarded work instead; do not use for wars or battles \\ 
P1411 & nominated for & award nomination received by a person, organisation or creative work (inspired from "award received" (Property:P166)) \\ 

\hline
\end{tabular}
}
\label{table:relation-info}
\end{table*}

\section{PAT-SND Model Implementation Details}
\label{appendix-sec:gat-implementation}

In our experiments, BERT\footnote{We use the BERT model ``bert-base-cased" as text encoder. We expect that using larger transformer embedding leads to better results. But due to our limitation of computational resources, we only did experiments based on this base BERT model.}~\cite{Devlin2019bert} is used to produce text embedding. To produce BERT embedding, the input of BERT is formatted by adding ``[CLS]" before and ``[SEP]" after the tokens of the description. This input is tokenized by the BERT tokenizer into word pieces. The output of the pretrained BERT model embedding is a sequence of vectors, each of size 768. Each output vector corresponds to one word piece token. BERT tokenizer tokenizes some words into word pieces (sub-word tokens), such as ``tokenizer" is tokenized as word pieces ``token" and ``\#\#izer". We take the average of the word pieces embedding of the original word to obtain the embedding of this word.

We empirically set PAT-SND hyper-parameters as follows: 

\begin{itemize}
    \item The method of choosing hyperparameter values is based on manual tuning to find the best AUC score. 
    \item The hidden state size as 300D; BERT embeddings mapped into 300D using a linear layer.
    \item There are 8 attention heads used for the PAT layers.
    \item The mini-batch size is set as 256. We use larger batch size to make training process faster. We searched the batch sizes in set \{32, 64, 128, 256\}.
    \item The learning rate is set as 0.001, searched in the set \{5e-5, 1e-4, 5e-4, 1e-3\}.
    \item We apply $l_2$ regularization with term $\lambda=10^{-4}$.
    \item Adam~\cite{kingma2014adam} optimizer is used for training. 
    \item Training runtime: The model is trained with 10 epochs. Each epoch takes around 60 minutes to run.
    \item Inference runtime: The inference time for 2000 test instances is 0.4 minute.
    \item The number of parameters of PAT-SND is 1,902,360.
\end{itemize}

The implementation of this model is based on PyTorch and NVIDIA GPU GTX 2080 Ti.

\section{Data Annotation Guideline\footnote{This annotation guideline is written for our volunteer annotators during the data annotation process. We include it in appendix of this paper.}}
\label{appendix-sec:annoation-guideline}

\begin{figure}
\centering
\scalebox{0.75}{
\begin{tabular}{cp{6.8cm}|l}
\vspace*{-3mm}
$d_1$~: & ``\textit{\underline{Iron Man} is a 2008 American superhero film based on the Marvel Comics character of the same name, stars \underline{Robert Downey Jr.} et al.}" & \textbf{normal}   \\    \\


$d_2$~: & ``\textit{In the 2010 Marvel film \underline{Iron Man 2}, \underline{Elon Musk} appeared in a scene with Tony Stark as a rival/friend.}'' & \textbf{novel} \\  \\

$d_3$~: & ``\textit{Austrian-American actress \underline{Hedy Lamarr} is the co-inventor of an early technique for \underline{Frequency-hopping spread spectrum} }.'' &  \textbf{novel} \\ 

\end{tabular}}
\caption{Examples of semantic novelty detection in factual texts involving named entities (underlined).}
\label{fig:annotation_exp}
\end{figure}

\subsection{Semantic Novelty Detection Involving Named Entities Annotation Goals}
This paper proposes the new task - Semantic Novelty Detection in Factual Text Involving Named Entities. Given a factual text $d$ containing two named entities, The goal is to classify whether a given factual text $d$ represents a semantically novel fact or a normal one with respect to the entity pair.

For instance, as shown in Figure~\ref{fig:annotation_exp}, the entity pairs $d_1$ and $d_2$ have the same relation ``cast member'' (predefined in a Knowledge Repository (KR)). $d_1$ is a normal fact with respect to the underlined name entities, because it is natural for an actor (Robert Downey Jr.) to act in a film (Iron Man). However, $d_2$ is a novel fact with respect to the underlined pair of entities because a CEO of a technology company (Elon Musk) acting in a film (Iron Man 2) is very novel and surprising.

In this annotation task, we focus on 20 human-related relations (see details in Table~\ref{table:relation-info}) as the annotation of novel facts related to these relations does not require extensive domain knowledge. 

For each relation $r$, our goal is to annotate 50 novel instances. Each instance is a text $d$ with two entities. These two entities semantically express the relation based on the contextual information in the text. 

\subsection{What is Semantic Novelty in This Task?}
The semantic novelty for a factual text involving named entities is that the two named entities have a novel interaction in the text that violates some common sense. For instance, it is commonsense that (c1) - ``an actor is a cast member of a movie'', (c2) - ``a scientist invented a technological device''. The factual text violates the commonsense knowledge is a semantically novel factual text. For instance, $d_2$ is semantically novel because it violates (c1). $d_3$ is semantically novel because it violates (c2). 

Note that, semantic novelty is subjective and personal. It happens that a factual text may be novel to one annotator but not others. In this work, we restrict our study to the consensus-view of semantic novelty. That is, a majority of people agree that the instance is novel. Thus, the annotators vote whether or not an annotated instance is novel and select the novel instances that a majority of the annotators agree. 

\subsection{Annotation Format}

Annotators are free to write a factual text from scratch or paraphrase from existing ones from online resource such as blogs, news articles. The final annotation format is shown in Table~\ref{table:annotation-format}, which shows the one novel instance in XML format. The meanings of the tags are self-explanatory. Briefly, each instance is defined as an ``instance'' element, which contains two named entity elements ``e1'' and ``e2''. Each named entity pair has a label, a description (optional) and a property value list. In the ``property\_value'' tag, each property value pair is a list with a property id (e.g., P31), a property label (e.g., instance of) and value (e.g., television series, ...), separated by a separator ``$||$''. 

Note that, the named entities annotated in the test data are not required to be chosen from the existing ones in the Knowledge Repository (KR) - Wikidata. The annotators are free to choose either of the two options: (1) use existing named entities from the KR. In this case, a python script is provided to the annotators to output the property-value pairs of the named entity in KR. (2) create a new named entity from scratch based on his/her knowledge, as long as its properties (expressed as property ids) are contained in Wikidata.

\begin{table*}
\centering
\caption{Annotation Format}
\scalebox{0.8}{
\begin{tabular}{p{18cm}}
\hline

<instance> \\
<instance\_id>1</instance\_id> \\
<text> Despite his status and very busy schedule, <e1>\textcolor{blue}{Elon Musk}</e1> still performs as an guest actor in <e2>\textcolor{blue}{The Big Bang Theory}</e2> as himself, surprising fellow engineer, Howard, by working along with him in a soup kitchen. </text> \\
\\
<e1> \\
<id>Q8539</id> \\
<label>The Big Bang Theory</label> \\
<description>American television sitcom 2007-2019</description> \\
<property\_value> \\
P31 || instance of || television series, connected set of television program episodes under the same title \\
P57 || director || Mark Cendrowski, American television director \\
P58 || screenwriter || ["Chuck Lorre, American televison director, screenwriter, producer, composer and actor", "Bill Prady, American television writer and producer", "Steven Molaro, Television producer and writer"]  \\
P136 || genre || American television sitcom, television sitcom series originating from the USA \\
</property\_value> \\
</e1> \\

\\

<e2> \\
<id>Q317521</id> \\
<label>Elon Musk</label> \\
<description>business magnate (born 1971)</description> \\
<property\_value> \\
P19 || place of birth || Pretoria, administrative capital of South Africa located in the Gauteng province  \\
P21 || sex or gender || male, to be used in "sex or gender" (P21) to indicate that the human subject is a male \\
P22 || father || Errol Musk, South African electromechanical engineer  \\
P25 || mother || Maye Musk, Canadian-born American model and dietitian \\
P26 || spouse || ["Justine Musk, Canadian writer", "Talulah Riley, British actress", "Talulah Riley, British actress"] \\
P27 || country of citizenship || ["South Africa, sovereign state in Southern Africa", "Canada, sovereign state in North America", "United States of America, sovereign state in North America"] \\
P31 || instance of || human, common name of Homo sapiens, unique extant species of the genus Homo \\
P40 || child || ["Griffin Musk,", "Xavier Musk,", "Damian Musk,", "Saxon Musk,", "Kai Musk,", "X \u00c6 A-XII Musk, child of Grimes and Elon Musk"]  \\
P106 || occupation || ["inventor, person that devises a new device, method, composition, or process", "programmer, person who writes computer software", "engineer, professional practitioner of engineering and its sub classes", "entrepreneur, individual who organizes and operates a business, taking on financial risk to do so"] \\
</property\_value> \\
</e2> \\

</instance> \\

\hline
\end{tabular}
}
\label{table:annotation-format}
\end{table*}

\section{Attention Illustration of PAT-SND Model for 20 relations}
\label{appendix-sec:att-illustration}
\begin{table*}[t!]
\centering
\captionsetup{font=scriptsize}
\caption{PAT-SND attention illustration for relation P6, P39's normal and novel entity pairs.}
\scalebox{0.65}{
\begin{tabular}{ C{0.3cm} | p{0.6cm} p{4cm} p{5cm} | p{0.6cm} p{4cm} p{5cm} }
\hline
L & \multicolumn{6}{c}{Text and two entities for relation P6: head of government  } \\ 
\hline
\multirow{12}{0.3cm}{\textbf{R}}& \multicolumn{6}{p{19.2cm}}{Following the AKP 's landslide victory in 2002 , the party 's co - founder \colorbox{blue!20}{Abdullah Gül} became Prime Minister , until \colorbox{green!20}{his government} annulled Erdoğan 's ban from political office .} \\
& \multicolumn{3}{c|}{\cellcolor{green!20}Cabinet Gül} & \multicolumn{3}{c}{\cellcolor{blue!20}Abdullah Gül} \\
& 10.54 & instance of & cabinet & 5.43 & description & 11th President of Turkey  \\
& 9.94 & country & Turkey & 4.49 & instance of & human  \\
& 9.94 & followed by & Cabinet Erdoğan I & 3.91 & position held & President of Turkey  \\
& 9.94 & start time & time +2002-01-01T00:00:00Z timezone 0 before 0 ... & 3.90 & occupation & politician  \\
\cline{2-7}
& \multicolumn{6}{p{19.2cm}}{The mayor of \colorbox{green!20}{Copenhagen} , \colorbox{blue!20}{Frank Jensen} , declared in late August that the city would contribute to the budget with 40 million ( Danish Kroner ) ( ) .} \\
& \multicolumn{3}{c|}{\cellcolor{green!20}Copenhagen Municipality} & \multicolumn{3}{c}{\cellcolor{blue!20}Frank Jensen} \\
& 6.65 & description & municipality in the Capital Region of Denmark & 7.72 & description & Danish politician  \\
& 5.47 & instance of & municipality of Denmark & 6.40 & instance of & human  \\
& 5.28 & Commons category & Københavns Kommune & 5.58 & position held & Justice Minister of Denmark  \\
& 5.17 & social media followers & amount +11038 unit 1 & 5.56 & occupation & politician  \\
\cline{1-7}
\multirow{12}{0.3cm}{\textbf{V}} & \multicolumn{6}{p{19.2cm}}{\colorbox{blue!20}{Volodymyr Oleksandrovych Zelenskyy} serves as the sixth and current president of \colorbox{green!20}{Ukraine} since 2019.} \\
& \multicolumn{3}{c|}{\cellcolor{green!20}Ukraine} & \multicolumn{3}{c}{\cellcolor{blue!20}Volodymyr Zelenskyy} \\
& 1.35 & description & country in Eastern Europe & 4.09 & description & sixth and current President of Ukraine  \\
& 1.10 & instance of & sovereign state & 3.36 & instance of & human  \\
& 1.06 & Commons category & Ukraine & 2.92 & occupation & screenwriter  \\
& 1.04 & described by source & Brockhaus and Efron Encyclopedic Dictionary & 2.91 & Commons category & Volodymyr Zelenskyy  \\
\cline{2-7}
& \multicolumn{6}{p{19.2cm}}{\colorbox{blue!20}{Ronald Wilson Reagan} was an American politician who served as the 40th president of the \colorbox{green!20}{United States} from 1981 to 1989 .} \\
& \multicolumn{3}{c|}{\cellcolor{green!20}United States of America} & \multicolumn{3}{c}{\cellcolor{blue!20}Ronald Reagan} \\
& 1.19 & description & country located mainly in North America & 2.42 & description & president of the United States from 1981 to 1989  \\
& 0.97 & instance of & sovereign state & 1.98 & instance of & human  \\
& 0.93 & Commons category & United States & 1.73 & position held & Governor of California  \\
& 0.92 & described by source & Small Brockhaus and Efron Encyclopedic Dictionary & 1.72 & occupation & television actor  \\
\hline
L & \multicolumn{6}{c}{Text and two entities for relation P39: position held  } \\ 
\hline
\multirow{12}{0.3cm}{\textbf{R}}& \multicolumn{6}{p{19.2cm}}{Born in Nashua , New Hampshire , he is the son of Catherine Gregg ( née Warner ) and \colorbox{green!20}{Hugh Gregg} , who was \colorbox{blue!20}{Governor} from 1953 to 1955 .} \\
& \multicolumn{3}{c|}{\cellcolor{green!20}Hugh Gregg} & \multicolumn{3}{c}{\cellcolor{blue!20}Governor of New Hampshire} \\
& 6.43 & occupation & politician & 7.27 & description & head of state and of government of the U.S. st ...  \\
& 6.39 & description & American politician (1917-2003) & 6.05 & instance of & elective office  \\
& 6.19 & instance of & human & 5.79 & label & Governor of New Hampshire  \\
& 5.83 & member of political party & Republican Party & 5.79 & topic's main category & Category:Governors of New Hampshire  \\
\cline{2-7}
& \multicolumn{6}{p{19.2cm}}{The carvings are possibly the arms of \colorbox{green!20}{William Booth} , \colorbox{blue!20}{Bishop of Lichfield} .} \\
& \multicolumn{3}{c|}{\cellcolor{green!20}William Booth} & \multicolumn{3}{c}{\cellcolor{blue!20}Bishop of Lichfield} \\
& 7.28 & occupation & Catholic priest & 11.10 & description & diocesan bishop in the Church of England  \\
& 7.23 & description & Archbishop of York & 9.25 & instance of & position  \\
& 7.01 & instance of & human & 8.87 & label & Bishop of Lichfield  \\
& 6.56 & described by source & Dictionary of National Biography & 8.87 & topic's main category & Category:Bishops of Lichfield  \\
\cline{1-7}
\multirow{12}{0.3cm}{\textbf{V}} & \multicolumn{6}{p{19.2cm}}{The construction bill is signed by the \colorbox{blue!20}{Governor of California} \colorbox{green!20}{Arnold Schwarzenegger} .} \\
& \multicolumn{3}{c|}{\cellcolor{green!20}Arnold Schwarzenegger} & \multicolumn{3}{c}{\cellcolor{blue!20}Governor of California} \\
& 2.30 & occupation & actor & 5.91 & description & head of government in the US state of California  \\
& 2.29 & description & Austrian-American actor & 4.91 & instance of & elective office  \\
& 2.21 & instance of & human & 4.71 & spouse & First Lady or Partner of California  \\
& 2.09 & award received & Grand Gold Decoration of Styria & 4.70 & label & Governor of California  \\
\cline{2-7}
& \multicolumn{6}{p{19.2cm}}{\colorbox{green!20}{Zelenskyy} as the \colorbox{blue!20}{President of Ukraine} condemns 'deliberate Russian war crime' after POW bombing .} \\
& \multicolumn{3}{c|}{\cellcolor{green!20}Volodymyr Zelenskyy} & \multicolumn{3}{c}{\cellcolor{blue!20}President of Ukraine} \\
& 2.34 & occupation & screenwriter & 7.27 & description & head of state of Ukraine  \\
& 2.33 & description & sixth and current President of Ukraine & 6.04 & instance of & position  \\
& 2.26 & instance of & human & 5.79 & label & President of Ukraine  \\
& 2.20 & instrument & voice & 5.79 & topic's main category & Category:Presidency of Ukraine  \\
\hline
\end{tabular}
}
\label{table:relation-examples-0}
\end{table*}


\begin{table*}[t!]
\centering
\captionsetup{font=scriptsize}
\caption{PAT-SND attention illustration for relation P57, P58's normal and novel entity pairs.}
\scalebox{0.65}{
\begin{tabular}{ C{0.3cm} | p{0.6cm} p{4cm} p{5cm} | p{0.6cm} p{4cm} p{5cm} }
\hline
L & \multicolumn{6}{c}{Text and two entities for relation P57: director  } \\ 
\hline
\multirow{12}{0.3cm}{\textbf{R}}& \multicolumn{6}{p{19.2cm}}{He was also associated with the film \colorbox{green!20}{Chaturanga} as an Chief AD directed by \colorbox{blue!20}{Suman Mukhopadhya} , participated in Montréal World Film Festival .} \\
& \multicolumn{3}{c|}{\cellcolor{green!20}Chaturanga} & \multicolumn{3}{c}{\cellcolor{blue!20}Suman Mukhopadhyay} \\
& 12.06 & description & 2008 film by Suman Mukhopadhyay & 6.51 & occupation & film director  \\
& 11.92 & instance of & film & 6.19 & instance of & human  \\
& 10.87 & composer & Debojyoti Mishra & 6.13 & description & Indian film director  \\
& 10.87 & cast member & Rituparna Sengupta & 5.88 & related category & Category:Films directed by Suman Mukhopadhyay  \\
\cline{2-7}
& \multicolumn{6}{p{19.2cm}}{In 1942 the novel was used as the basis for the historical film " \colorbox{green!20}{Luisa Sanfelice} " directed by \colorbox{blue!20}{Leo Menardi} .} \\
& \multicolumn{3}{c|}{\cellcolor{green!20}Luisa Sanfelice} & \multicolumn{3}{c}{\cellcolor{blue!20}Leo Menardi} \\
& 5.79 & description & 1942 Italian historical drama film directed by ... & 7.88 & occupation & film director  \\
& 5.72 & instance of & film & 7.49 & instance of & human  \\
& 5.23 & genre & drama & 7.42 & description & Italian screenwriter and film director  \\
& 5.21 & composer & Renzo Rossellini & 7.12 & related category & Category:Films directed by Leo Menardi  \\
\cline{1-7}
\multirow{12}{0.3cm}{\textbf{V}} & \multicolumn{6}{p{19.2cm}}{\colorbox{green!20}{Secret} is a 2007 Taiwanese film directed by Taiwanese \colorbox{blue!20}{Jay Chou} .} \\
& \multicolumn{3}{c|}{\cellcolor{green!20}Secret} & \multicolumn{3}{c}{\cellcolor{blue!20}Jay Chou} \\
& 6.10 & description & 2007 film by Jay Chou & 3.60 & occupation & actor  \\
& 6.03 & instance of & film & 3.42 & instance of & human  \\
& 5.51 & genre & musical film & 3.39 & description & Taiwanese musician  \\
& 5.50 & different from & Secret & 3.27 & instrument & piano  \\
\cline{2-7}
& \multicolumn{6}{p{19.2cm}}{\colorbox{green!20}{Piranha II: The Spawning} is a 1982 American independent horror film directed by \colorbox{blue!20}{James Cameron} in his feature directorial debut.} \\
& \multicolumn{3}{c|}{\cellcolor{green!20}Piranha II: The Spawning} & \multicolumn{3}{c}{\cellcolor{blue!20}James Cameron} \\
& 5.01 & description & 1981 film by James Cameron & 3.84 & occupation & janitor  \\
& 4.95 & instance of & film & 3.65 & instance of & human  \\
& 4.52 & genre & horror film & 3.61 & description & Canadian filmmaker  \\
& 4.51 & composer & Stelvio Cipriani & 3.47 & award received & Academy Award for Best Director  \\
\hline
L & \multicolumn{6}{c}{Text and two entities for relation P58: screenwriter  } \\ 
\hline
\multirow{12}{0.3cm}{\textbf{R}}& \multicolumn{6}{p{19.2cm}}{Eastwood and Siegel hired a new writer , \colorbox{blue!20}{Dean Riesner} , who had written for Siegel in the Henry Fonda TV film " \colorbox{green!20}{Stranger on the Run} " .} \\
& \multicolumn{3}{c|}{\cellcolor{green!20}Stranger on the Run} & \multicolumn{3}{c}{\cellcolor{blue!20}Dean Riesner} \\
& 9.72 & description & 1967 television film directed by Don Siegel & 7.13 & occupation & screenwriter  \\
& 8.52 & instance of & television film & 7.11 & instance of & human  \\
& 7.48 & cast member & Henry Fonda & 6.74 & description & American screenwriter (1918-2002)  \\
& 7.48 & genre & Western film & 6.70 & Commons category & Dean Riesner  \\
\cline{2-7}
& \multicolumn{6}{p{19.2cm}}{\colorbox{blue!20}{John Requa} is an American screenwriter ( with Glenn Ficarra ) of " Cats \& Dogs " , " Bad Santa " and the 2005 remake " \colorbox{green!20}{Bad News Bears} " .} \\
& \multicolumn{3}{c|}{\cellcolor{green!20}Bad News Bears} & \multicolumn{3}{c}{\cellcolor{blue!20}John Requa} \\
& 5.59 & description & 2005 film by Richard Linklater & 8.21 & occupation & screenwriter  \\
& 4.89 & instance of & film & 8.18 & instance of & human  \\
& 4.29 & genre & comedy film & 7.75 & description & American writer  \\
& 4.29 & cast member & Billy Bob Thornton & 7.71 & Commons category & John Requa  \\
\cline{1-7}
\multirow{12}{0.3cm}{\textbf{V}} & \multicolumn{6}{p{19.2cm}}{\colorbox{blue!20}{Jay Chou} is one of the screenwriter of the 2007 Taiwanese film \colorbox{green!20}{Secret} .} \\
& \multicolumn{3}{c|}{\cellcolor{green!20}Secret} & \multicolumn{3}{c}{\cellcolor{blue!20}Jay Chou} \\
& 7.10 & description & 2007 film by Jay Chou & 3.47 & occupation & actor  \\
& 6.22 & instance of & film & 3.46 & instance of & human  \\
& 5.46 & cast member & Gwei Lun-Mei & 3.28 & description & Taiwanese musician  \\
& 5.46 & genre & musical film & 3.26 & Commons category & Jay Chou  \\
\cline{2-7}
& \multicolumn{6}{p{19.2cm}}{\colorbox{green!20}{Renaldo and Clara} is a 1978 American film written by \colorbox{blue!20}{Bob Dylan} and Sam Shepard .} \\
& \multicolumn{3}{c|}{\cellcolor{green!20}Renaldo and Clara} & \multicolumn{3}{c}{\cellcolor{blue!20}Bob Dylan} \\
& 12.50 & description & 1978 film by Bob Dylan & 2.04 & occupation & songwriter  \\
& 10.97 & instance of & film & 2.03 & instance of & human  \\
& 9.64 & genre & drama & 1.94 & related category & Category:Films directed by Bob Dylan  \\
& 9.56 & color & color & 1.92 & description & American singer-songwriter (born 1941)  \\
\hline
\end{tabular}
}
\label{table:relation-examples-1}
\end{table*}


\begin{table*}[t!]
\centering
\captionsetup{font=scriptsize}
\caption{PAT-SND attention illustration for relation P61, P84's normal and novel entity pairs.}
\scalebox{0.65}{
\begin{tabular}{ C{0.3cm} | p{0.6cm} p{4cm} p{5cm} | p{0.6cm} p{4cm} p{5cm} }
\hline
L & \multicolumn{6}{c}{Text and two entities for relation P61: discoverer or inventor  } \\ 
\hline
\multirow{12}{0.3cm}{\textbf{R}}& \multicolumn{6}{p{19.2cm}}{Working with the noted Australian astrophotographer \colorbox{blue!20}{David Malin} , they discovered the largest spiral galaxy known , dubbed \colorbox{green!20}{Malin 1} .} \\
& \multicolumn{3}{c|}{\cellcolor{green!20}Malin 1} & \multicolumn{3}{c}{\cellcolor{blue!20}David Malin} \\
& 8.61 & description & low-surface-brightness spiral galaxy & 8.03 & occupation & astronomer  \\
& 6.13 & instance of & low-surface-brightness galaxy & 7.82 & instance of & human  \\
& 6.10 & Commons category & Malin 1 & 7.49 & description & British-Australian astronomical photographer  \\
& 6.09 & constellation & Coma Berenices & 7.08 & award received & Jackson-Gwilt Medal  \\
\cline{2-7}
& \multicolumn{6}{p{19.2cm}}{While these lamps are now antiques , the technology of the neon glow lamp developed into contemporary plasma displays and televisions . \colorbox{green!20}{Neon} was discovered in 1898 by the British scientists \colorbox{blue!20}{William Ramsay} and Morris W. Travers .} \\
& \multicolumn{3}{c|}{\cellcolor{green!20}neon} & \multicolumn{3}{c}{\cellcolor{blue!20}William Ramsay} \\
& 6.32 & description & chemical element with symbol Ne and atomic num ... & 3.57 & occupation & chemist  \\
& 4.49 & instance of & chemical element & 3.47 & instance of & human  \\
& 4.46 & Commons category & Neon & 3.33 & description & Scottish Chemist  \\
& 4.46 & part of & period 2 & 3.21 & Commons category & William Ramsay  \\
\cline{1-7}
\multirow{12}{0.3cm}{\textbf{V}} & \multicolumn{6}{p{19.2cm}}{\colorbox{blue!20}{Hedy Lamarr} is the co-inventor of an early technique for \colorbox{green!20}{spread spectrum communications and frequency hopping.}} \\
& \multicolumn{3}{c|}{\cellcolor{green!20}Frequency-hopping spread spectrum} & \multicolumn{3}{c}{\cellcolor{blue!20}Hedy Lamarr} \\
& 31.41 & description & radio signal transmission method & 3.02 & occupation & actor  \\
& 22.96 & instance of & technique & 2.93 & instance of & human  \\
& 22.84 & subclass of & spread spectrum & 2.81 & description & Austrian-American actress  \\
& 22.79 & label & Frequency-hopping spread spectrum & 2.71 & Commons category & Hedy Lamarr  \\
\cline{2-7}
& \multicolumn{6}{p{19.2cm}}{\colorbox{blue!20}{Forence Lawrence} invented the predecessor of the \colorbox{green!20}{auto signaling arm} .} \\
& \multicolumn{3}{c|}{\cellcolor{green!20}automotive lighting} & \multicolumn{3}{c}{\cellcolor{blue!20}Florence Lawrence} \\
& 25.73 & description & lighting system of a motor vehicle & 4.95 & occupation & actor  \\
& 18.59 & Commons category & Automobile lights & 4.81 & instance of & human  \\
& 18.58 & topic's main category & Category:Automotive lamps & 4.61 & description & Canadian-American actress (1886-1938)  \\
& 18.57 & subclass of & light source & 4.45 & Commons category & Florence Lawrence  \\
\hline
L & \multicolumn{6}{c}{Text and two entities for relation P84: architect  } \\ 
\hline
\multirow{12}{0.3cm}{\textbf{R}}& \multicolumn{6}{p{19.2cm}}{The building 's façade closely resembled the \colorbox{blue!20}{Bradford Gilbert} - designed \colorbox{green!20}{Illinois Central Station} in Chicago that had opened in 1893 .} \\
& \multicolumn{3}{c|}{\cellcolor{green!20}Central Station} & \multicolumn{3}{c}{\cellcolor{blue!20}Bradford Gilbert} \\
& 10.79 & instance of & railway station & 8.55 & description & American architect  \\
& 10.69 & description & railroad terminal in Chicago & 8.48 & occupation & architect  \\
& 9.94 & Commons category & Central Station (Chicago) & 7.81 & instance of & human  \\
& 9.85 & label & Central Station & 7.56 & sex or gender & male  \\
\cline{2-7}
& \multicolumn{6}{p{19.2cm}}{Due to the split , Lyon moved into the \colorbox{green!20}{Stade de Gerland} , a stadium designed by local architect \colorbox{blue!20}{Tony Garnier} .} \\
& \multicolumn{3}{c|}{\cellcolor{green!20}Stade de Gerland} & \multicolumn{3}{c}{\cellcolor{blue!20}Tony Garnier} \\
& 5.75 & instance of & multi-purpose stadium & 4.50 & description & French architect  \\
& 5.69 & description & stadium in Lyon & 4.46 & occupation & architect  \\
& 5.28 & Commons category & Stade de Gerland & 4.10 & instance of & human  \\
& 5.23 & label & Stade de Gerland & 4.00 & award received & Prix de Rome  \\
\cline{1-7}
\multirow{12}{0.3cm}{\textbf{V}} & \multicolumn{6}{p{19.2cm}}{The Chapel of \colorbox{green!20}{Exeter College , Oxford} , designed by \colorbox{blue!20}{James Rooke} , was consecrated by the Bishop of Oxford on St Luke 's Day 1859 .} \\
& \multicolumn{3}{c|}{\cellcolor{green!20}Exeter College} & \multicolumn{3}{c}{\cellcolor{blue!20}James Rooke} \\
& 6.42 & instance of & college of the University of Oxford & 7.41 & description & English general in the British Army  \\
& 6.35 & description & constituent college of the University of Oxford & 7.34 & occupation & politician  \\
& 5.90 & Commons category & Exeter College & 6.77 & position held & Member of the 1st Parliament of the United Kingdom  \\
& 5.84 & label & Exeter College & 6.76 & instance of & human  \\
\cline{2-7}
& \multicolumn{6}{p{19.2cm}}{This masque was the first one performed in the new \colorbox{green!20}{Banqueting House} in Whitehall Palace , designed and built by \colorbox{blue!20}{Filippo Trenta} after the previous wooden structure burned down in January 1619 .} \\
& \multicolumn{3}{c|}{\cellcolor{green!20}Banqueting House} & \multicolumn{3}{c}{\cellcolor{blue!20}Filippo Trenta} \\
& 6.42 & instance of & banqueting house & 9.22 & description & roman-catholic bishop  \\
& 6.35 & description & former palace banqueting rooms & 9.13 & occupation & Catholic priest  \\
& 5.90 & Commons category & Banqueting House & 8.43 & position held & Catholic bishop  \\
& 5.84 & label & Banqueting House & 8.42 & instance of & human  \\
\hline
\end{tabular}
}
\label{table:relation-examples-2}
\end{table*}


\begin{table*}[t!]
\centering
\captionsetup{font=scriptsize}
\caption{PAT-SND attention illustration for relation P86, P161's normal and novel entity pairs.}
\scalebox{0.65}{
\begin{tabular}{ C{0.3cm} | p{0.6cm} p{4cm} p{5cm} | p{0.6cm} p{4cm} p{5cm} }
\hline
L & \multicolumn{6}{c}{Text and two entities for relation P86: composer  } \\ 
\hline
\multirow{12}{0.3cm}{\textbf{R}}& \multicolumn{6}{p{19.2cm}}{His greatest operatic success was in the leading role in " \colorbox{green!20}{Peter Grimes} " , an opera by \colorbox{blue!20}{Benjamin Britten} .} \\
& \multicolumn{3}{c|}{\cellcolor{green!20}Peter Grimes} & \multicolumn{3}{c}{\cellcolor{blue!20}Benjamin Britten} \\
& 12.05 & description & opera by Benjamin Britten & 3.15 & description & English composer  \\
& 11.36 & instance of & opera & 2.49 & position held & Member of the House of Lords  \\
& 11.05 & Commons category & Peter Grimes & 2.49 & occupation & conductor  \\
& 10.94 & language of work or name & English & 2.48 & instance of & human  \\
\cline{2-7}
& \multicolumn{6}{p{19.2cm}}{The book takes its name from a Donna Summer cover of the song \colorbox{green!20}{" MacArthur Park "} , originally sung by Richard Harris and written / composed by \colorbox{blue!20}{Jimmy Webb} .} \\
& \multicolumn{3}{c|}{\cellcolor{green!20}MacArthur Park} & \multicolumn{3}{c}{\cellcolor{blue!20}Jimmy Webb} \\
& 10.87 & description & original song written and composed by Jimmy We ... & 4.42 & description & American songwriter  \\
& 10.25 & instance of & musical composition & 3.51 & occupation & singer-songwriter  \\
& 9.87 & language of work or name & English & 3.49 & instance of & human  \\
& 9.87 & title & MacArthur Park & 3.45 & instrument & piano  \\
\cline{1-7}
\multirow{12}{0.3cm}{\textbf{V}} & \multicolumn{6}{p{19.2cm}}{The Swiss entry was Céline Dion with the French language song " \colorbox{green!20}{Ne partez pas sans moi} " ( Do n't leave without me ) , composed by \colorbox{blue!20}{William Morgan} and Nella Martinetti .} \\
& \multicolumn{3}{c|}{\cellcolor{green!20}Ne partez pas sans moi} & \multicolumn{3}{c}{\cellcolor{blue!20}William Morgan} \\
& 7.80 & description & 1988 Céline Dion song & 9.66 & description & Welsh Jesuit  \\
& 7.34 & instance of & single & 7.71 & occupation & college head  \\
& 7.16 & genre & pop music & 7.68 & instance of & human  \\
& 7.09 & record label & Columbia Records & 7.52 & sex or gender & male  \\
\cline{2-7}
& \multicolumn{6}{p{19.2cm}}{He believed that \colorbox{blue!20}{Antonio Donghi} 's " \colorbox{green!20}{Salome} " ( 1905 ) was the most important work of recent modern music .} \\
& \multicolumn{3}{c|}{\cellcolor{green!20}Salome} & \multicolumn{3}{c}{\cellcolor{blue!20}Antonio Donghi} \\
& 6.43 & description & opera by Richard Strauss & 6.69 & description & Italian painter (1897-1963)  \\
& 6.06 & instance of & dramatico-musical work & 5.32 & occupation & painter  \\
& 5.90 & discography & Salome discography & 5.30 & instance of & human  \\
& 5.89 & Commons category & Salome (opera) & 5.20 & Commons category & Antonio Donghi  \\
\hline
L & \multicolumn{6}{c}{Text and two entities for relation P161: cast member  } \\ 
\hline
\multirow{12}{0.3cm}{\textbf{R}}& \multicolumn{6}{p{19.2cm}}{In 2009 , he starred in `` He 's Just Not That Into You '' along with co-star Ginnifer Goodwin and `` \colorbox{green!20}{After.Life} '' opposite \colorbox{blue!20}{Liam Neeson} and Christina Ricci .} \\
& \multicolumn{3}{c|}{\cellcolor{green!20}After.Life} & \multicolumn{3}{c}{\cellcolor{blue!20}Liam Neeson} \\
& 5.83 & description & 2009 psychological horror-thriller film by Agn ... & 3.32 & occupation & film actor  \\
& 3.88 & instance of & film & 3.29 & description & Northern Irish actor  \\
& 3.65 & genre & horror film & 3.12 & position held & UNICEF Goodwill Ambassador  \\
& 3.61 & director & Agnieszka Wojtowicz-Vosloo & 3.12 & instance of & human  \\
\cline{2-7}
& \multicolumn{6}{p{19.2cm}}{He starred alongside \colorbox{blue!20}{Chris Kattan} in the film `` \colorbox{green!20}{Christmas in Wonderland} '' .} \\
& \multicolumn{3}{c|}{\cellcolor{green!20}Christmas in Wonderland} & \multicolumn{3}{c}{\cellcolor{blue!20}Chris Kattan} \\
& 9.59 & description & 2007 film by James Orr & 6.81 & occupation & screenwriter  \\
& 6.44 & instance of & television film & 6.73 & description & American actor and comedian  \\
& 6.06 & genre & children's film & 6.39 & instance of & human  \\
& 6.00 & director & James Orr & 6.18 & Commons category & Chris Kattan  \\
\cline{1-7}
\multirow{12}{0.3cm}{\textbf{V}} & \multicolumn{6}{p{19.2cm}}{Despite his status and very busy schedule, \colorbox{blue!20}{Elon Musk} still performs as an guest actor in \colorbox{green!20}{The Big Bang Theory} as himself, surprising fellow engineer, Howard, by working along with him in a soup kitchen.} \\
& \multicolumn{3}{c|}{\cellcolor{green!20}The Big Bang Theory} & \multicolumn{3}{c}{\cellcolor{blue!20}Elon Musk} \\
& 3.89 & description & American television sitcom 2007-2019 & 2.56 & occupation & inventor  \\
& 2.57 & instance of & television series & 2.53 & description & business magnate (born 1971)  \\
& 2.47 & has part or parts & The Big Bang Theory & 2.40 & position held & chief executive officer  \\
& 2.46 & Commons category & The Big Bang Theory & 2.40 & instance of & human  \\
\cline{2-7}
& \multicolumn{6}{p{19.2cm}}{\colorbox{blue!20}{Jeff Bezos} was unrecognizable in the 2016 sci-fi film \colorbox{green!20}{Star Trek Beyond} and his eight-second cameo proved to be even more challenging to notice due to the heavy prosthetics and makeup that he sported.} \\
& \multicolumn{3}{c|}{\cellcolor{green!20}Star Trek Beyond} & \multicolumn{3}{c}{\cellcolor{blue!20}Jeff Bezos} \\
& 2.73 & description & 2016 film directed by Justin Lin & 2.75 & occupation & computer scientist  \\
& 1.80 & instance of & 3D film & 2.72 & description & American engineer and entrepreneur  \\
& 1.72 & Commons category & Star Trek Beyond & 2.58 & position held & chief executive officer  \\
& 1.69 & genre & science fiction film & 2.58 & instance of & human  \\
\hline
\end{tabular}
}
\label{table:relation-examples-3}
\end{table*}


\begin{table*}[t!]
\centering
\captionsetup{font=scriptsize}
\caption{PAT-SND attention illustration for relation P170, P175's normal and novel entity pairs.}
\scalebox{0.65}{
\begin{tabular}{ C{0.3cm} | p{0.6cm} p{4cm} p{5cm} | p{0.6cm} p{4cm} p{5cm} }
\hline
L & \multicolumn{6}{c}{Text and two entities for relation P170: creator  } \\ 
\hline
\multirow{12}{0.3cm}{\textbf{R}}& \multicolumn{6}{p{19.2cm}}{Time to Hunt is a 1999 thriller novel , and the third in the \colorbox{green!20}{Bob Lee Swagger} series by \colorbox{blue!20}{Stephen Hunter} .} \\
& \multicolumn{3}{c|}{\cellcolor{green!20}Bob Lee Swagger} & \multicolumn{3}{c}{\cellcolor{blue!20}Stephen Hunter} \\
& 8.64 & description & fictional United States Marine & 5.60 & instance of & human  \\
& 7.41 & instance of & fictional human & 5.44 & occupation & film critic  \\
& 7.41 & occupation & soldier & 5.41 & description & American novelist  \\
& 7.00 & sex or gender & male & 5.34 & sex or gender & male  \\
\cline{2-7}
& \multicolumn{6}{p{19.2cm}}{In 2001 , he worked with \colorbox{blue!20}{Victoria Pile} on a new series `` \colorbox{green!20}{Los Dos Bros} '' , an off-beat sitcom exploring physical comedy and the relationship between Boyd and Cavan Clerkin as the titular ( half- ) brothers .} \\
& \multicolumn{3}{c|}{\cellcolor{green!20}Los Dos Bros} & \multicolumn{3}{c}{\cellcolor{blue!20}Victoria Pile} \\
& 12.04 & description & television series & 10.55 & instance of & human  \\
& 10.32 & instance of & television series & 10.26 & occupation & television director  \\
& 9.76 & genre & sitcom & 10.19 & description & British television director and producer  \\
& 9.70 & number of seasons & amount +1 unit 1 & 10.06 & sex or gender & female  \\
\cline{1-7}
\multirow{12}{0.3cm}{\textbf{V}} & \multicolumn{6}{p{19.2cm}}{\colorbox{blue!20}{Jordi Branes} cast her as Stephanie Tanner in the ABC comedy series `` \colorbox{green!20}{Full House} '' in 1987 , and she played that role until the show ended in 1995 .} \\
& \multicolumn{3}{c|}{\cellcolor{green!20}Full House} & \multicolumn{3}{c}{\cellcolor{blue!20}Jordi Branes} \\
& 4.91 & description & American sitcom television series & 20.85 & instance of & human  \\
& 4.20 & instance of & television series & 20.32 & occupation & military personnel  \\
& 4.09 & has part or parts & Full House & 19.92 & sex or gender & male  \\
& 3.97 & genre & American television sitcom & 19.46 & label & Jordi Branes  \\
\cline{2-7}
& \multicolumn{6}{p{19.2cm}}{The plot of \colorbox{blue!20}{Mason Jones} 's 2004 \colorbox{green!20}{Harry Bosch} novel , `` The Narrows '' , revolves around a crime committed in Zzyzx .} \\
& \multicolumn{3}{c|}{\cellcolor{green!20}Harry Bosch} & \multicolumn{3}{c}{\cellcolor{blue!20}Mason Jones} \\
& 7.15 & description & Fictional detective created by author Michael  ... & 5.90 & instance of & human  \\
& 6.13 & instance of & fictional human & 5.74 & occupation & basketball player  \\
& 6.13 & occupation & soldier & 5.70 & description & American basketball player  \\
& 5.79 & sex or gender & male & 5.62 & sex or gender & male  \\
\hline
L & \multicolumn{6}{c}{Text and two entities for relation P175: performer  } \\ 
\hline
\multirow{12}{0.3cm}{\textbf{R}}& \multicolumn{6}{p{19.2cm}}{In 2014 , Shiroyan decided to take part in season four of " The Voice of Ukraine " , auditioning with the Polish song " \colorbox{green!20}{Dziwny jest ten świat} " by \colorbox{blue!20}{Czesław Niemen} .} \\
& \multicolumn{3}{c|}{\cellcolor{green!20}Dziwny jest ten świat…} & \multicolumn{3}{c}{\cellcolor{blue!20}Czesław Niemen} \\
& 8.61 & instance of & album & 4.63 & description & Polish rock musician  \\
& 8.02 & description & 1967 debut studio album by Czesław Niemen \& A ... & 3.99 & occupation & composer  \\
& 7.62 & genre & soul music & 3.74 & instance of & human  \\
& 7.58 & language of work or name & Polish & 3.64 & instrument & organ  \\
\cline{2-7}
& \multicolumn{6}{p{19.2cm}}{" \colorbox{green!20}{Rough Day} " is a song by Australian recording artist \colorbox{blue!20}{Paulini} , taken from her second studio album , " Superwoman " ( 2006 ) .} \\
& \multicolumn{3}{c|}{\cellcolor{green!20}Rough Day} & \multicolumn{3}{c}{\cellcolor{blue!20}Paulini} \\
& 18.32 & instance of & single & 8.41 & description & Australian singer  \\
& 17.07 & description & 2006 single by Paulini & 7.29 & occupation & singer  \\
& 16.24 & genre & pop music & 6.86 & instance of & human  \\
& 16.15 & publication date & time +2006-01-22T00:00:00Z timezone 0 before 0 ... & 6.68 & instrument & voice  \\
\cline{1-7}
\multirow{12}{0.3cm}{\textbf{V}} & \multicolumn{6}{p{19.2cm}}{She was the narrator for \colorbox{blue!20}{Liliana Barańska} 's 1971 experimental jazz composition " \colorbox{green!20}{Escalator over the Hill} " .} \\
& \multicolumn{3}{c|}{\cellcolor{green!20}Escalator over the Hill} & \multicolumn{3}{c}{\cellcolor{blue!20}Liliana Barańska} \\
& 9.07 & description & album & 8.40 & description & Polish politician  \\
& 8.55 & instance of & album & 6.71 & member of political party & Democratic Left Alliance  \\
& 8.33 & genre & avant-garde jazz & 6.69 & occupation & politician  \\
& 8.26 & record label & Jazz Composer's Orchestra & 6.66 & instance of & human  \\
\cline{2-7}
& \multicolumn{6}{p{19.2cm}}{" \colorbox{green!20}{Emotional} " is a 1986 song by Austrian pop musician \colorbox{blue!20}{Thomas Harlow} from his album " \colorbox{green!20}{Emotional} " .} \\
& \multicolumn{3}{c|}{\cellcolor{green!20}Emotional} & \multicolumn{3}{c}{\cellcolor{blue!20}Thomas Harlow} \\
& 8.61 & instance of & album & 11.29 & description & college basketball player (1952–1952) Massachu ...  \\
& 8.03 & description & album by Falco & 9.82 & occupation & basketball player  \\
& 7.62 & genre & pop rock & 9.27 & instance of & human  \\
& 7.58 & language of work or name & German & 8.82 & member of sports team & UMass Minutemen basketball  \\
\hline
\end{tabular}
}
\label{table:relation-examples-4}
\end{table*}


\begin{table*}[t!]
\centering
\captionsetup{font=scriptsize}
\caption{PAT-SND attention illustration for relation P241, P412's normal and novel entity pairs.}
\scalebox{0.65}{
\begin{tabular}{ C{0.3cm} | p{0.6cm} p{4cm} p{5cm} | p{0.6cm} p{4cm} p{5cm} }
\hline
L & \multicolumn{6}{c}{Text and two entities for relation P241: military branch  } \\ 
\hline
\multirow{12}{0.3cm}{\textbf{R}}& \multicolumn{6}{p{19.2cm}}{\colorbox{green!20}{Albert Cushing Read} ( 1887–1967 ) was an aviator and admiral in the \colorbox{blue!20}{United States Navy} .} \\
& \multicolumn{3}{c|}{\cellcolor{green!20}Albert Cushing Read} & \multicolumn{3}{c}{\cellcolor{blue!20}United States Navy} \\
& 6.93 & description & United States Navy admiral and aviator & 6.53 & description & maritime warfare branch of the United States'  ...  \\
& 5.82 & occupation & military officer & 5.96 & instance of & navy  \\
& 5.55 & instance of & human & 5.65 & conflict & American Revolutionary War  \\
& 5.20 & Commons category & Albert Cushing Read & 5.59 & Commons category & United States Navy  \\
\cline{2-7}
& \multicolumn{6}{p{19.2cm}}{Major Levison \colorbox{green!20}{James Wood} ( born 5 May 1982 ) is a \colorbox{blue!20}{British Army} officer and explorer .} \\
& \multicolumn{3}{c|}{\cellcolor{green!20}Sir James Wood, 2nd Baronet} & \multicolumn{3}{c}{\cellcolor{blue!20}British Army} \\
& 14.04 & description & British Army general & 6.55 & description & principal land warfare force of the United Kingdom  \\
& 11.89 & occupation & military leader & 5.97 & instance of & army  \\
& 11.33 & instance of & human & 5.66 & conflict & World War I  \\
& 10.58 & military rank & general & 5.60 & Commons category & British Army  \\
\cline{1-7}
\multirow{12}{0.3cm}{\textbf{V}} & \multicolumn{6}{p{19.2cm}}{This changed on 16 May 1855 when \colorbox{green!20}{Alfonso Maria Giordano} assumed command of the \colorbox{blue!20}{French Army} , and agreed with Lord Raglan that the Russian fortifications should be assaulted .} \\
& \multicolumn{3}{c|}{\cellcolor{green!20}Alfonso Maria Giordano} & \multicolumn{3}{c}{\cellcolor{blue!20}French Army} \\
& 7.98 & occupation & Catholic priest & 9.69 & description & land warfare branch of France's military  \\
& 7.63 & instance of & human & 8.85 & instance of & army  \\
& 7.41 & position held & Catholic archbishop & 8.40 & conflict & World War I  \\
& 7.11 & sex or gender & male & 8.31 & Commons category & French Army  \\
\cline{2-7}
& \multicolumn{6}{p{19.2cm}}{The \colorbox{blue!20}{British Army} commander , Major General \colorbox{green!20}{Kent Twitchell} , was killed in the same action .} \\
& \multicolumn{3}{c|}{\cellcolor{green!20}Kent Twitchell} & \multicolumn{3}{c}{\cellcolor{blue!20}British Army} \\
& 11.65 & description & American artist & 6.55 & description & principal land warfare force of the United Kingdom  \\
& 9.83 & occupation & painter & 5.97 & instance of & army  \\
& 9.37 & instance of & human & 5.66 & conflict & World War I  \\
& 8.79 & Commons category & Kent Twitchell & 5.60 & Commons category & British Army  \\
\hline
L & \multicolumn{6}{c}{Text and two entities for relation P412: voice type  } \\ 
\hline
\multirow{12}{0.3cm}{\textbf{R}}& \multicolumn{6}{p{19.2cm}}{\colorbox{green!20}{Josephine Veasey} ( born 10 July 1930 ) is a British \colorbox{blue!20}{mezzo - soprano} , particularly associated with Wagner and Berlioz 
 roles .} \\
& \multicolumn{3}{c|}{\cellcolor{green!20}Josephine Veasey} & \multicolumn{3}{c}{\cellcolor{blue!20}mezzo-soprano} \\
& 9.25 & description & singer & 12.77 & description & type of classical female singing voice whose v ...  \\
& 8.65 & instance of & human & 11.57 & instance of & voice type  \\
& 8.56 & occupation & singer & 10.97 & Commons category & Mezzo-sopranos  \\
& 8.28 & instrument & voice & 10.90 & instrument & voice  \\
\cline{2-7}
& \multicolumn{6}{p{19.2cm}}{\colorbox{green!20}{Éric Huchet} ( born in 1952 in Saint - Germain - en - Laye ) is a French contemporary lyric \colorbox{blue!20}{tenor} .} \\
& \multicolumn{3}{c|}{\cellcolor{green!20}Eric Huchet} & \multicolumn{3}{c}{\cellcolor{blue!20}tenor} \\
& 9.24 & description & French singer & 12.77 & description & classical male singing voice  \\
& 8.64 & instance of & human & 11.57 & instance of & voice type  \\
& 8.55 & occupation & singer & 10.97 & Commons category & Tenors  \\
& 8.28 & instrument & voice & 10.90 & instrument & voice  \\
\cline{1-7}
\multirow{12}{0.3cm}{\textbf{V}} & \multicolumn{6}{p{19.2cm}}{Her cousin \colorbox{green!20}{Svetlana Evgenevna Ermijaeva} was a famous \colorbox{blue!20}{soprano} .} \\
& \multicolumn{3}{c|}{\cellcolor{green!20}Svetlana Evgenevna Ermijaeva} & \multicolumn{3}{c}{\cellcolor{blue!20}soprano} \\
& 13.07 & instance of & human & 11.54 & description & type of classical female singing voice  \\
& 12.93 & occupation & visual artist & 10.45 & instance of & profession  \\
& 12.46 & sex or gender & female & 9.91 & Commons category & Soprano vocalists  \\
& 12.32 & country of citizenship & Russia & 9.84 & instrument & voice  \\
\cline{2-7}
& \multicolumn{6}{p{19.2cm}}{Among her pupils was British \colorbox{blue!20}{soprano} \colorbox{green!20}{Aura Castro} .} \\
& \multicolumn{3}{c|}{\cellcolor{green!20}Aura Castro} & \multicolumn{3}{c}{\cellcolor{blue!20}soprano} \\
& 8.56 & description & Chilean sculptress & 11.54 & description & type of classical female singing voice  \\
& 8.00 & instance of & human & 10.45 & instance of & profession  \\
& 7.92 & occupation & sculptor & 9.91 & Commons category & Soprano vocalists  \\
& 7.63 & sex or gender & female & 9.84 & instrument & voice  \\
\hline
\end{tabular}
}
\label{table:relation-examples-5}
\end{table*}


\begin{table*}[t!]
\centering
\captionsetup{font=scriptsize}
\caption{PAT-SND attention illustration for relation P413, P463's normal and novel entity pairs.}
\scalebox{0.65}{
\begin{tabular}{ C{0.3cm} | p{0.6cm} p{4cm} p{5cm} | p{0.6cm} p{4cm} p{5cm} }
\hline
L & \multicolumn{6}{c}{Text and two entities for relation P413: position played on team / speciality  } \\ 
\hline
\multirow{12}{0.3cm}{\textbf{R}}& \multicolumn{6}{p{19.2cm}}{The Vikings defense ranked sixth in the league in points allowed and was led by Hall of Fame \colorbox{blue!20}{defensive tackle} \colorbox{green!20}{John Randle} .} \\
& \multicolumn{3}{c|}{\cellcolor{green!20}John Randle} & \multicolumn{3}{c}{\cellcolor{blue!20}defensive tackle} \\
& 7.10 & description & player of American football & 18.35 & description & position in American football  \\
& 7.07 & occupation & American football player & 16.53 & instance of & American football position  \\
& 6.98 & instance of & human & 16.42 & sport & American football  \\
& 6.67 & sex or gender & male & 16.25 & label & defensive tackle  \\
\cline{2-7}
& \multicolumn{6}{p{19.2cm}}{\colorbox{green!20}{Louis Linwood Voit} ( born February 13 , 1991 ) is an American professional baseball \colorbox{blue!20}{first baseman} for the St. Louis Cardinals of Major League Baseball ( MLB ) .} \\
& \multicolumn{3}{c|}{\cellcolor{green!20}Luke Voit} & \multicolumn{3}{c}{\cellcolor{blue!20}first baseman} \\
& 6.66 & description & Professional baseball player & 12.36 & description & defensive position in baseball and softball  \\
& 6.63 & occupation & baseball player & 11.11 & instance of & baseball position  \\
& 6.54 & instance of & human & 11.04 & sport & baseball  \\
& 6.28 & Commons category & Luke Voit & 10.93 & location & first base  \\
\cline{1-7}
\multirow{12}{0.3cm}{\textbf{V}} & \multicolumn{6}{p{19.2cm}}{\colorbox{green!20}{Adrianus Valerius} is a Danish football \colorbox{blue!20}{defender} who currently plays for Middelfart Boldklub in the Danish 2nd Division .} \\
& \multicolumn{3}{c|}{\cellcolor{green!20}Adrianus Valerius} & \multicolumn{3}{c}{\cellcolor{blue!20}defender} \\
& 5.34 & description & Dutch National Anthem writer & 10.15 & description & sports position played near the player's team' ...  \\
& 5.31 & occupation & poet & 9.12 & instance of & association football position  \\
& 5.25 & instance of & human & 9.06 & sport & association football  \\
& 5.04 & Commons category & Adriaen Valerius & 8.98 & part of & defense  \\
\cline{2-7}
& \multicolumn{6}{p{19.2cm}}{\colorbox{green!20}{Jean-baptiste Ernest Boulage} is a Canadian former ice hockey \colorbox{blue!20}{right winger} .} \\
& \multicolumn{3}{c|}{\cellcolor{green!20}Jean-baptiste Ernest Boulage} & \multicolumn{3}{c}{\cellcolor{blue!20}winger} \\
& 10.57 & description & French official (1807-1863) & 12.36 & description & ice hockey position  \\
& 10.52 & occupation & official & 11.11 & instance of & ice hockey position  \\
& 10.39 & instance of & human & 11.04 & sport & ice hockey  \\
& 9.93 & sex or gender & male & 10.92 & subclass of & forward  \\
\hline
L & \multicolumn{6}{c}{Text and two entities for relation P463: member of  } \\ 
\hline
\multirow{12}{0.3cm}{\textbf{R}}& \multicolumn{6}{p{19.2cm}}{The album featured a guest appearance from \colorbox{green!20}{Simone Simons} of \colorbox{blue!20}{Epica} , who also appeared on " Gods of Vermin " .} \\
& \multicolumn{3}{c|}{\cellcolor{green!20}Simone Simons} & \multicolumn{3}{c}{\cellcolor{blue!20}Epica} \\
& 4.79 & occupation & singer & 7.36 & description & Dutch symphonic metal band  \\
& 4.61 & instance of & human & 5.56 & Commons category & Epica  \\
& 4.60 & description & Dutch singer & 5.51 & instance of & musical group  \\
& 4.57 & genre & symphonic metal & 5.47 & discography & Epica discography  \\
\cline{2-7}
& \multicolumn{6}{p{19.2cm}}{\colorbox{green!20}{David Hurn} ( born 21 July 1934 ) is a British documentary photographer and member of \colorbox{blue!20}{Magnum Photos} .} \\
& \multicolumn{3}{c|}{\cellcolor{green!20}David Hurn} & \multicolumn{3}{c}{\cellcolor{blue!20}Magnum Photos} \\
& 6.57 & occupation & photographer & 10.07 & description & international photographic cooperative  \\
& 6.32 & instance of & human & 7.64 & Commons category & Magnum Photos  \\
& 6.30 & description & British photographer & 7.57 & instance of & business  \\
& 6.27 & Commons category & David Hurn & 7.48 & label & Magnum Photos  \\
\cline{1-7}
\multirow{12}{0.3cm}{\textbf{V}} & \multicolumn{6}{p{19.2cm}}{The last song recorded in the 1982 sessions was the country soul ballad " Love Bankrupt " , written by \colorbox{green!20}{Theodor Rutt} and Linda Womack of \colorbox{blue!20}{Womack \& Womack} .} \\
& \multicolumn{3}{c|}{\cellcolor{green!20}Theodor Rutt} & \multicolumn{3}{c}{\cellcolor{blue!20}Womack \& Womack} \\
& 7.50 & occupation & university teacher & 15.95 & description & American musical duo  \\
& 7.22 & award received & Order of Merit of North Rhine-Westphalia & 12.30 & has part or parts & Cecil Womack  \\
& 7.21 & instance of & human & 12.08 & instance of & musical duo  \\
& 7.20 & description & German university teacher and writer (1911-2006) & 11.99 & discography & Womack \& Womack discography  \\
\cline{2-7}
& \multicolumn{6}{p{19.2cm}}{\colorbox{green!20}{Ángel Puig Puig} also mentioned that it 'll be hard to keep \colorbox{blue!20}{Nevermore} legacy alive , since Jeff Loomis will be tough to replace .} \\
& \multicolumn{3}{c|}{\cellcolor{green!20}Ángel Puig Puig} & \multicolumn{3}{c}{\cellcolor{blue!20}Nevermore} \\
& 8.70 & occupation & politician & 7.76 & description & American heavy metal band  \\
& 8.51 & member of political party & Autonomist Republican Union Party & 5.92 & has part or parts & Warrel Dane  \\
& 8.39 & position held & Member of the Cortes republicanas & 5.87 & Commons category & Nevermore  \\
& 8.38 & instance of & human & 5.81 & instance of & musical group  \\
\hline
\end{tabular}
}
\label{table:relation-examples-6}
\end{table*}


\begin{table*}[t!]
\centering
\captionsetup{font=scriptsize}
\caption{PAT-SND attention illustration for relation P641, P800's normal and novel entity pairs.}
\scalebox{0.65}{
\begin{tabular}{ C{0.3cm} | p{0.6cm} p{4cm} p{5cm} | p{0.6cm} p{4cm} p{5cm} }
\hline
L & \multicolumn{6}{c}{Text and two entities for relation P641: sport  } \\ 
\hline
\multirow{12}{0.3cm}{\textbf{R}}& \multicolumn{6}{p{19.2cm}}{\colorbox{green!20}{Jamila Wideman} ( born October 16 , 1975 ) is an American female left - handed point guard \colorbox{blue!20}{basketball} player , lawyer , and activist .} \\
& \multicolumn{3}{c|}{\cellcolor{green!20}Jamila Wideman} & \multicolumn{3}{c}{\cellcolor{blue!20}basketball} \\
& 7.51 & instance of & human & 4.47 & instance of & type of sport  \\
& 7.45 & description & American basketball player & 4.42 & Commons category & Basketball  \\
& 6.55 & given name & Jamila & 4.38 & description & team sport played on a court with baskets on e ...  \\
& 6.55 & country of citizenship & United States of America & 4.37 & topic's main category & Category:Basketball  \\
\cline{2-7}
& \multicolumn{6}{p{19.2cm}}{\colorbox{green!20}{Martha Nelson} ( born 22 October 1954 ) is a Canadian former \colorbox{blue!20}{swimmer} .} \\
& \multicolumn{3}{c|}{\cellcolor{green!20}Martha Nelson} & \multicolumn{3}{c}{\cellcolor{blue!20}swimming} \\
& 11.15 & instance of & human & 8.54 & instance of & type of sport  \\
& 11.06 & description & Canadian swimmer & 8.44 & Commons category & Competitive swimming  \\
& 9.74 & given name & Martha & 8.37 & description & water-based sport  \\
& 9.74 & country of citizenship & Canada & 8.36 & topic's main category & Category:Swimming  \\
\cline{1-7}
\multirow{12}{0.3cm}{\textbf{V}} & \multicolumn{6}{p{19.2cm}}{Pavel Svojanovský ( born 12 August 1943 ) is a retired Czech \colorbox{blue!20}{rower} who mostly competed in the coxed pairs , together with his younger brother \colorbox{green!20}{Yosuke Sakai} .} \\
& \multicolumn{3}{c|}{\cellcolor{green!20}Yosuke Sakai} & \multicolumn{3}{c}{\cellcolor{blue!20}rowing} \\
& 13.79 & instance of & human & 6.84 & instance of & Olympic sport  \\
& 13.68 & description & Japanese designer & 6.76 & Commons category & Rowing  \\
& 12.38 & occupation & designer & 6.70 & description & sport where individuals or teams row boats by oar  \\
& 12.05 & country of citizenship & Japan & 6.70 & topic's main category & Category:Rowing  \\
\cline{2-7}
& \multicolumn{6}{p{19.2cm}}{\colorbox{green!20}{Emil Murray} is a French male \colorbox{blue!20}{volleyball} player .} \\
& \multicolumn{3}{c|}{\cellcolor{green!20}Emil Murray} & \multicolumn{3}{c}{\cellcolor{blue!20}volleyball} \\
& 18.56 & instance of & human & 4.89 & instance of & type of sport  \\
& 16.67 & occupation & opinion journalist & 4.85 & has part or parts & volleyball rules  \\
& 16.23 & given name & Emil & 4.83 & Commons category & Volleyball  \\
& 16.23 & label & Emil Murray & 4.79 & description & ballgame and team sport in which two teams com ...  \\
\hline
L & \multicolumn{6}{c}{Text and two entities for relation P800: notable work  } \\ 
\hline
\multirow{12}{0.3cm}{\textbf{R}}& \multicolumn{6}{p{19.2cm}}{It also includes a homage to \colorbox{green!20}{Larry Niven} 's " \colorbox{blue!20}{Ringworld} " ( 1970).Larry Niven , " N - Space " , pp .} \\
& \multicolumn{3}{c|}{\cellcolor{green!20}Larry Niven} & \multicolumn{3}{c}{\cellcolor{blue!20}Ringworld} \\
& 5.57 & description & American writer & 9.83 & description & 1970 Larry Niven science fiction novel  \\
& 4.78 & sex or gender & male & 9.04 & instance of & written work  \\
& 4.75 & occupation & writer & 7.47 & award received & Hugo Award for Best Novel  \\
& 4.75 & award received & Inkpot Award & 7.44 & genre & science fiction novel  \\
\cline{2-7}
& \multicolumn{6}{p{19.2cm}}{The episode was directed by former " \colorbox{blue!20}{Breaking Bad} " writer \colorbox{green!20}{John Shiban} .} \\
& \multicolumn{3}{c|}{\cellcolor{green!20}John Shiban} & \multicolumn{3}{c}{\cellcolor{blue!20}Breaking Bad} \\
& 11.57 & description & American television writer and producer & 3.88 & description & American television series (2008–2013)  \\
& 9.94 & sex or gender & male & 3.54 & instance of & television series  \\
& 9.89 & occupation & screenwriter & 3.02 & Commons category & Breaking Bad  \\
& 9.84 & nominated for & Primetime Emmy Award for Outstanding Writing f ... & 2.95 & has part or parts & Breaking Bad  \\
\cline{1-7}
\multirow{12}{0.3cm}{\textbf{V}} & \multicolumn{6}{p{19.2cm}}{In 1847 , Goffredo Mameli and \colorbox{green!20}{Abdallah Ben Barek} composed " \colorbox{blue!20}{Il Canto degli Italiani} " .} \\
& \multicolumn{3}{c|}{\cellcolor{green!20}Abdallah Ben Barek} & \multicolumn{3}{c}{\cellcolor{blue!20}Il Canto degli Italiani} \\
& 7.77 & description & Moroccan association football player & 8.57 & description & national anthem of Italy  \\
& 6.67 & sex or gender & male & 7.86 & instance of & song  \\
& 6.63 & occupation & association football player & 6.71 & Commons category & Il Canto degli Italiani  \\
& 6.61 & member of sports team & Granada CF & 6.42 & part of & National symbols of Italy  \\
\cline{2-7}
& \multicolumn{6}{p{19.2cm}}{Based on \colorbox{green!20}{Aykut Emre Yakut} 's musical \colorbox{blue!20}{of the same name} , the film is written and directed by Richard Lagravenese .} \\
& \multicolumn{3}{c|}{\cellcolor{green!20}Aykut Emre Yakut} & \multicolumn{3}{c}{\cellcolor{blue!20}The Last Five Years} \\
& 11.56 & description & Turkish association football player & 36.80 & description & 2001 musical  \\
& 9.94 & sex or gender & male & 34.83 & instance of & musical theatre  \\
& 9.89 & occupation & association football player & 28.38 & label & The Last Five Years  \\
&   &   &   &   &   &    \\
\hline
\end{tabular}
}
\label{table:relation-examples-7}
\end{table*}


\begin{table*}[t!]
\centering
\captionsetup{font=scriptsize}
\caption{PAT-SND attention illustration for relation P991, P1303's normal and novel entity pairs.}
\scalebox{0.65}{
\begin{tabular}{ C{0.3cm} | p{0.6cm} p{4cm} p{5cm} | p{0.6cm} p{4cm} p{5cm} }
\hline
L & \multicolumn{6}{c}{Text and two entities for relation P991: successful candidate  } \\ 
\hline
\multirow{12}{0.3cm}{\textbf{R}}& \multicolumn{6}{p{19.2cm}}{Baron was a strong backer of David Davis in the 2005 Conservative leadership election , having also supported him in the \colorbox{green!20}{2001 leadership contest} won by \colorbox{blue!20}{Iain Duncan Smith} .} \\
& \multicolumn{3}{c|}{\cellcolor{green!20}2001 Conservative Party (UK) leadership election} & \multicolumn{3}{c}{\cellcolor{blue!20}Iain Duncan Smith} \\
& 14.01 & description & British Conservative Party leadership election & 3.93 & description & British politician  \\
& 13.24 & instance of & leadership election & 3.74 & occupation & politician  \\
& 12.13 & office contested & Leader of the Conservative Party & 3.73 & instance of & human  \\
& 12.13 & point in time & time +2001-00-00T00:00:00Z timezone 0 before 0 ... & 3.38 & Commons category & Iain Duncan Smith  \\
\cline{2-7}
& \multicolumn{6}{p{19.2cm}}{Caldwell lost to former Honolulu Prosecuting Attorney \colorbox{blue!20}{Peter Carlisle} in the \colorbox{green!20}{2010 special Mayoral election} .} \\
& \multicolumn{3}{c|}{\cellcolor{green!20}2010 Honolulu mayoral election} & \multicolumn{3}{c}{\cellcolor{blue!20}Peter Carlisle} \\
& 15.38 & instance of & mayoral election & 7.90 & description & politician  \\
& 14.36 & Commons category & Honolulu mayoral special election & 7.54 & occupation & politician  \\
& 14.06 & office contested & mayor & 7.53 & instance of & human  \\
& 14.06 & point in time & time +2010-00-00T00:00:00Z timezone 0 before 0 ... & 6.82 & Commons category & Peter Carlisle  \\
\cline{1-7}
\multirow{12}{0.3cm}{\textbf{V}} & \multicolumn{6}{p{19.2cm}}{On October 7, 2003, actor \colorbox{blue!20}{Arnold Schwarzenegger} is elected governor of California, in \colorbox{green!20}{a special recall election} to replace then-Governor Gray Davis.} \\
& \multicolumn{3}{c|}{\cellcolor{green!20}2003 California gubernatorial recall election} & \multicolumn{3}{c}{\cellcolor{blue!20}Arnold Schwarzenegger} \\
& 11.27 & description & Special election for the governorship of the U ... & 2.51 & description & Austrian-American actor  \\
& 10.64 & instance of & gubernatorial election & 2.39 & occupation & actor  \\
& 9.94 & Commons category & California gubernatorial recall election & 2.38 & instance of & human  \\
& 9.74 & office contested & Governor of California & 2.16 & Commons category & Arnold Schwarzenegger  \\
\cline{2-7}
& \multicolumn{6}{p{19.2cm}}{Late - arriving evidence included a letter dated 17 December 1992 from \colorbox{blue!20}{William F. Ruddiman} , who had become President of Iran after winning the \colorbox{green!20}{Iranian presidential election , 1980} .} \\
& \multicolumn{3}{c|}{\cellcolor{green!20}1980 Iranian presidential election} & \multicolumn{3}{c}{\cellcolor{blue!20}William F. Ruddiman} \\
& 11.27 & description & 1st Iranian presidential election & 7.05 & description & American palaeoclimatologist and professor  \\
& 10.64 & instance of & presidential election & 6.72 & occupation & geologist  \\
& 9.94 & Commons category & Iranian presidential election & 6.71 & instance of & human  \\
& 9.74 & office contested & President of Iran & 5.76 & award received & Lyell Medal  \\
\hline
L & \multicolumn{6}{c}{Text and two entities for relation P1303: instrument  } \\ 
\hline
\multirow{12}{0.3cm}{\textbf{R}}& \multicolumn{6}{p{19.2cm}}{The historic Walcker \colorbox{blue!20}{organ} has been used for recordings of music of the period , such as Martin Schmeding 's recording of the \colorbox{blue!20}{organ} works by \colorbox{green!20}{Max Reger} .} \\
& \multicolumn{3}{c|}{\cellcolor{green!20}Max Reger} & \multicolumn{3}{c}{\cellcolor{blue!20}organ} \\
& 4.51 & description & German composer & 11.99 & description & musical keyboard instrument  \\
& 4.23 & instance of & human & 8.24 & Commons category & Organs (music)  \\
& 3.79 & Commons category & Max Reger & 8.18 & has part or parts & organ case  \\
& 3.72 & sex or gender & male & 7.96 & described by source & Catholic Encyclopedia  \\
\cline{2-7}
& \multicolumn{6}{p{19.2cm}}{In Minor Threat , he originally played \colorbox{blue!20}{bass guitar} before switching to guitar in 1982 when \colorbox{green!20}{Steve Hansgen} joined the band , and then moved back to bass after Hansgen 's departure .} \\
& \multicolumn{3}{c|}{\cellcolor{green!20}Steve Hansgen} & \multicolumn{3}{c}{\cellcolor{blue!20}bass guitar} \\
& 16.37 & description & American musician & 14.23 & description & electric or acoustic bass instrument  \\
& 15.44 & instance of & human & 9.83 & Commons category & Bass guitars  \\
& 14.60 & occupation & musician & 9.50 & different from & electric bass guitar  \\
& 13.60 & sex or gender & male & 9.50 & used by & bass guitarist  \\
\cline{1-7}
\multirow{12}{0.3cm}{\textbf{V}} & \multicolumn{6}{p{19.2cm}}{This piece , particularly in a well - known arrangement for trumpet , string orchestra and \colorbox{blue!20}{organ} by Sir Henry Wood , was incorrectly attributed for years to \colorbox{green!20}{Charles O'Brien, 6th Viscount Clare} .} \\
& \multicolumn{3}{c|}{\cellcolor{green!20}Charles O'Brien, 6th Viscount Clare} & \multicolumn{3}{c}{\cellcolor{blue!20}organ} \\
& 6.04 & description & Jacobite noble & 11.99 & description & musical keyboard instrument  \\
& 5.66 & instance of & human & 8.24 & Commons category & Organs (music)  \\
& 5.35 & occupation & military personnel & 8.18 & has part or parts & organ case  \\
& 4.98 & sex or gender & male & 7.96 & described by source & Catholic Encyclopedia  \\
\cline{2-7}
& \multicolumn{6}{p{19.2cm}}{Allard studied \colorbox{blue!20}{clarinet} under \colorbox{green!20}{André Siewert} of the Boston Symphony and saxophone under Lyle Bowen .} \\
& \multicolumn{3}{c|}{\cellcolor{green!20}André Siewert} & \multicolumn{3}{c}{\cellcolor{blue!20}clarinet} \\
& 11.72 & description & German athlet & 13.01 & description & any unspecified or undetermined member of the  ...  \\
& 11.02 & instance of & human & 8.96 & Commons category & Clarinets  \\
& 10.42 & occupation & athletics competitor & 8.85 & award received & Instrument of the Year  \\
& 9.70 & sex or gender & male & 8.66 & described by source & Armenian Soviet Encyclopedia  \\
\hline
\end{tabular}
}
\label{table:relation-examples-8}
\end{table*}


\begin{table*}[t!]
\centering
\captionsetup{font=scriptsize}
\caption{PAT-SND attention illustration for relation P1346, P1411's normal and novel entity pairs.}
\scalebox{0.65}{
\begin{tabular}{ C{0.3cm} | p{0.6cm} p{4cm} p{5cm} | p{0.6cm} p{4cm} p{5cm} }
\hline
L & \multicolumn{6}{c}{Text and two entities for relation P1346: winner  } \\ 
\hline
\multirow{12}{0.3cm}{\textbf{R}}& \multicolumn{6}{p{19.2cm}}{The Penske PC4 was a Formula One car used by Team Penske during the 1976 and was driven to victory in that year 's \colorbox{green!20}{Austrian Grand Prix} by \colorbox{blue!20}{John Watson} .} \\
& \multicolumn{3}{c|}{\cellcolor{green!20}1976 Austrian Grand Prix} & \multicolumn{3}{c}{\cellcolor{blue!20}John Watson} \\
& 9.40 & instance of & Austrian Grand Prix & 6.73 & description & British racecar driver  \\
& 9.28 & description & 275th Formula 1 Championship Grand Prix & 5.72 & instance of & human  \\
& 8.16 & label & 1976 Austrian Grand Prix & 5.66 & Commons category & John Watson (racing driver)  \\
& 8.13 & point in time & time +1976-08-15T00:00:00Z timezone 0 before 0 ... & 5.61 & occupation & racing automobile driver  \\
\cline{2-7}
& \multicolumn{6}{p{19.2cm}}{The \colorbox{green!20}{2007 Championship} was won by \colorbox{blue!20}{John Higgins} who beat qualifier Mark Selby 18–13 in the final .} \\
& \multicolumn{3}{c|}{\cellcolor{green!20}2007 World Snooker Championship} & \multicolumn{3}{c}{\cellcolor{blue!20}John Higgins} \\
& 8.69 & instance of & snooker tournament & 5.78 & description & Scottish snooker player  \\
& 8.58 & description & snooker tournament & 4.91 & instance of & human  \\
& 7.54 & label & 2007 World Snooker Championship & 4.86 & Commons category & John Higgins  \\
& 7.52 & sponsor & 888 Holdings & 4.82 & occupation & snooker player  \\
\cline{1-7}
\multirow{12}{0.3cm}{\textbf{V}} & \multicolumn{6}{p{19.2cm}}{Winged Foot member \colorbox{blue!20}{Uldis Balodis} won three major titles : the 1927 U.S. Open , \colorbox{green!20}{1930 PGA Championship} , and the 1931 British Open .} \\
& \multicolumn{3}{c|}{\cellcolor{green!20}1930 PGA Championship} & \multicolumn{3}{c}{\cellcolor{blue!20}Uldis Balodis} \\
& 12.42 & instance of & PGA Championship & 11.56 & instance of & human  \\
& 12.26 & description & golf tournament held in 1930 & 11.34 & occupation & conductor  \\
& 10.80 & label & 1930 PGA Championship & 11.13 & sex or gender & male  \\
& 10.76 & coordinate location & latitude 40.733 longitude -73.78 altitude None ... & 11.01 & country of citizenship & Latvia  \\
\cline{2-7}
& \multicolumn{6}{p{19.2cm}}{Hotel du Lac is a 1984 \colorbox{green!20}{Booker Prize} - winning novel by English writer \colorbox{blue!20}{Lynn Paula} .} \\
& \multicolumn{3}{c|}{\cellcolor{green!20}Booker Prize} & \multicolumn{3}{c}{\cellcolor{blue!20}Lynn Paula} \\
& 8.06 & instance of & literary award & 11.94 & description & British actress  \\
& 7.96 & description & literary award & 10.19 & instance of & human  \\
& 7.19 & Commons category & Man Booker Prize & 9.98 & occupation & actor  \\
& 7.00 & label & Booker Prize & 9.80 & sex or gender & female  \\
\hline
L & \multicolumn{6}{c}{Text and two entities for relation P1411: nominated for  } \\ 
\hline
\multirow{12}{0.3cm}{\textbf{R}}& \multicolumn{6}{p{19.2cm}}{\colorbox{green!20}{Tony Gaudio} was nominated for the \colorbox{blue!20}{Academy Award for Best Black and White Cinematography} but lost to Gregg Toland for " Wuthering Heights " .} \\
& \multicolumn{3}{c|}{\cellcolor{green!20}Tony Gaudio} & \multicolumn{3}{c}{\cellcolor{blue!20}Academy Award for Best Cinematography} \\
& 7.43 & description & Italian American cinematographer & 15.99 & description & American film award  \\
& 6.18 & instance of & human & 12.71 & instance of & Academy Awards  \\
& 5.99 & occupation & cinematographer & 12.17 & Commons category & Academy Award for Best Cinematography  \\
& 5.98 & Commons category & Tony Gaudio & 11.84 & country & United States of America  \\
\cline{2-7}
& \multicolumn{6}{p{19.2cm}}{The reaction is named for \colorbox{blue!20}{Nobel Prize} winning chemist \colorbox{green!20}{Georg Wittig} .} \\
& \multicolumn{3}{c|}{\cellcolor{green!20}Georg Wittig} & \multicolumn{3}{c}{\cellcolor{blue!20}Nobel Prize in Chemistry} \\
& 5.54 & description & German chemist (1979 Nobel Prize) & 9.43 & description & one of the five Nobel Prizes established in 18 ...  \\
& 4.59 & instance of & human & 7.42 & instance of & science award  \\
& 4.45 & occupation & chemist & 7.11 & Commons category & Nobel Prize in Chemistry  \\
& 4.45 & Commons category & Georg Wittig & 6.93 & topic's main category & Category:Nobel Prize in Chemistry  \\
\cline{1-7}
\multirow{12}{0.3cm}{\textbf{V}} & \multicolumn{6}{p{19.2cm}}{At the Golden Raspberry Awards , the film was nominated for Worst Actress ( Miley Cyrus ) and \colorbox{blue!20}{Worst Supporting Actor} ( \colorbox{green!20}{Tyrell Lynch} ) .} \\
& \multicolumn{3}{c|}{\cellcolor{green!20}Tyrell Lynch} & \multicolumn{3}{c}{\cellcolor{blue!20}Golden Raspberry Award for Worst Supporting Actor} \\
& 10.40 & description & college basketball player (2009–2009) Massachu ... & 24.70 & description & award  \\
& 8.68 & instance of & human & 19.85 & instance of & Golden Raspberry Awards  \\
& 8.41 & occupation & basketball player & 18.50 & country & United States of America  \\
& 8.28 & sex or gender & male & 18.48 & inception & time +1981-00-00T00:00:00Z timezone 0 before 0 ...  \\
\cline{2-7}
& \multicolumn{6}{p{19.2cm}}{He is also nominated for the \colorbox{blue!20}{Academy Award for Best Production Design} for the film " Bridge of Spies " along with set decorators \colorbox{green!20}{Elia Meschak} and Rena DeAngelo .} \\
& \multicolumn{3}{c|}{\cellcolor{green!20}Elia Meschak} & \multicolumn{3}{c}{\cellcolor{blue!20}Academy Award for Best Production Design} \\
& 7.87 & description & Congolese association football player & 16.04 & description & Academy Award which recognizes achievement for ...  \\
& 6.55 & instance of & human & 12.75 & instance of & Academy Awards  \\
& 6.35 & occupation & association football player & 11.89 & topic's main category & Category:Best Art Direction Academy Award winners  \\
& 6.34 & Commons category & Meschack Elia & 11.88 & country & United States of America  \\
\hline
\end{tabular}
}
\label{table:relation-examples-9}
\end{table*}

Similar to the attention illustration of the PAT-SND model in Figure~\ref{fig:entity-pair-attention} (Sec.~\ref{sub-sec:novelty-characterization}) for relation ``discoverer/inventor'', we present the attention illustration for all 20 relations in this section from Table~\ref{table:relation-examples-0} to Table~\ref{table:relation-examples-9}. In these tables, we show two examples for both labels (L): NORMAL (R) and NOVEL (V). In the text, the two named entities are highlighted with different colors. For each named entity, we sort the property-value list in decreasing order based on the attention weights (represented as a percentage) and show the top 4 property-value pairs.

\section{Common Sense Knowledge}
\label{appendix-sec:common-sense-knowledge}

In this section, Table~\ref{table:common-sense-all-relation} presents the human summarized normal knowledge for all 20 relations. Because the 20 relations in our experiment are not domain-specific, the normal knowledge presented in Table~\ref{table:common-sense-all-relation} is common sense knowledge.

\begin{table*}[t!]
\centering
\caption{Common Sense Knowledge Summary of 20 Relations}
\scalebox{0.9}{
\begin{tabular}{C{1cm}|C{3cm}|p{10cm}}
\hline
ID & Label & \multicolumn{1}{c}{Common Sense} \\
\hline
P6 & head of government & The head of a government section is a person whose occupation is a politician. \\
P39 & position held & A position held by a person should be aligned with the occupation of this person. \\
P57 & director & A film (or a similar product) is directed by a director. \\
P58 & screenwriter & A film (or a similar product) is written by a screenwriter. \\
P61 & discoverer or inventor & A phenomenon/theory or an entity is discovered or invented by a person having an occupation in the same field. \\
P84 & architect & A building is designed by an architect. \\
P86 & composer & A musical composition (Opera or product with music related) is composed by a composer. \\
P161 & cast member & An actor is the cast member of a film (TV series or other similar product). \\
P170 & creator & A product is created by a person having an occupation in the same field. \\
P175 & performer & A musical work is performed by a musician or actor. \\
P241 & military branch & A person having an occupation related to the military is in a military branch. \\
P412 & voice type & A person with some voice type is a singer. \\
P413 & position played on team / speciality & A person's occupation aligns with the type of sports of the team in which this person plays a position. \\
P463 & member of & The field of the organization aligns with the occupation of the members. \\
P641 & sport & The type of sports aligns with the person's occupation. \\
P800 & notable work & The field of the notable work aligns with the creator's occupation field. \\
P991 & successful candidate & The successful candidate of an election is a politician. \\
P1303 & instrument & A person working in the music industry like a musician or a composer has an instrument. \\
P1346 & winner & The winner of a competition is a person having an occupation in the same field. \\
P1411 & nominated for & The nomination of the award is a person having an occupation in the same field. \\
\hline
\end{tabular}
}
\label{table:common-sense-all-relation}
\end{table*}


\end{document}